\pdfoutput=1

\documentclass[11pt]{article}

\usepackage[preprint]{acl}

\usepackage{times}
\usepackage{latexsym}

\usepackage[T1]{fontenc}

\usepackage[utf8]{inputenc}

\usepackage{microtype}

\usepackage{inconsolata}

\usepackage{graphicx}
\usepackage{hyperref}
\usepackage{booktabs}      
\usepackage{multirow}
\usepackage{array}
\usepackage{float}
\usepackage{amsmath}
\usepackage{amssymb}
\usepackage{bbm}
\usepackage{comment}
\usepackage[most]{tcolorbox}
\usepackage{enumitem}
\usepackage{algpseudocode}
\usepackage{algorithm}
\usepackage{subcaption}
\usepackage{arydshln}

\definecolor{darkgreen}{RGB}{0,140,0}

\newcommand{\method}{dynamic decomposition}
\newcommand{\short}{\textsc{DyDecomp}}

\title{
\vspace*{-0.5in}
{{\small \hfill ACL'25}\\
\vspace*{.25in}} 
Optimizing Decomposition for Optimal Claim Verification}

\author{Yining Lu \quad Noah Ziems \quad Hy Dang \quad Meng Jiang \\
University of Notre Dame \\ South Bend, IN \\
\texttt{ylu33@nd.edu}
}

\begin{document}
\maketitle
\begin{abstract}
Current research on the \textit{Decompose-Then-Verify} paradigm for evaluating the factuality of long-form text typically treats decomposition and verification in isolation, overlooking their interactions and potential misalignment.
We find that existing decomposition policies, typically hand-crafted demonstrations, do not align well with downstream verifiers in terms of atomicity\textemdash a novel metric quantifying information density\textemdash leading to suboptimal verification results.
We formulate finding the optimal decomposition policy for optimal verification as a bilevel optimization problem.
To approximate a solution for this strongly NP-hard problem, we propose \method{}, a reinforcement learning framework that leverages verifier feedback to learn a policy for dynamically decomposing claims to verifier-preferred atomicity.
Experimental results show that \method{} outperforms existing decomposition policies, improving verification confidence by 0.07 and accuracy by 0.12 (on a 0-1 scale) on average across varying verifiers, datasets, and atomcities of input claims.\footnote{Our code: \href{https://github.com/yining610/dynamic-decomposition}{github.com/yining610/dynamic-decomposition}}
\end{abstract}

\begin{figure*}
    \centering
    \includegraphics[width=\linewidth]{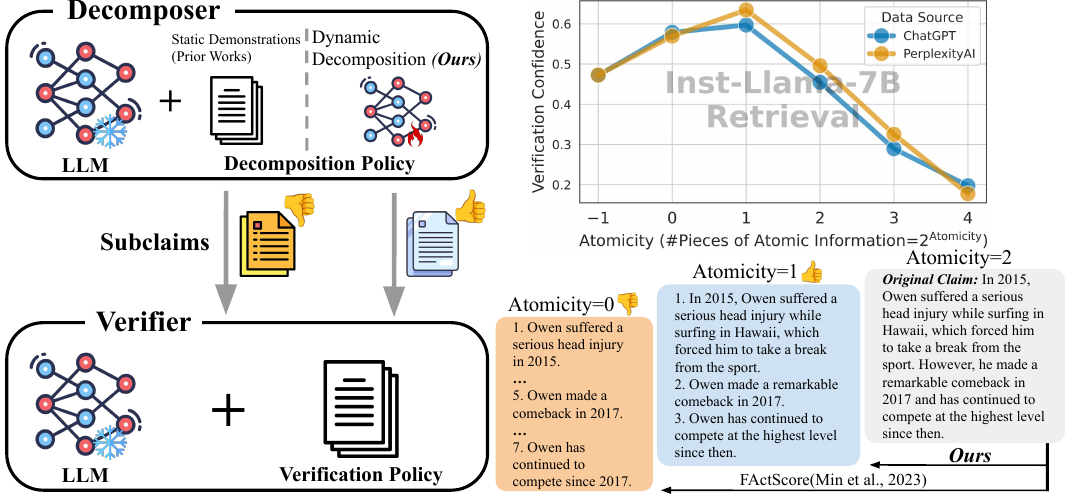}
    \caption{\textbf{Left}: overall framework of \textit{Decompose-Then-Verify} paradigm. We define each decomposer and verifier as a LLM paired with a corresponding policy. Our \method{} is compatible with existing fact-checking systems and requires training only a decomposition policy with 4.73M parameters. \textbf{Right}: the figure (\textit{upper right}) shows that the verification confidence of the verifier (i.e., Inst-Llama-7B with a retrieval verification policy) peaks at atomicity 1. An atomicity of -1 denotes the claim is partially trivial and tautological. The example (\textit{lower right}) shows that \textbf{the decomposition policy from FActScore \cite{min-etal-2023-factscore} fails to generate subclaims that best evoke the verifier's performance, leading to suboptimal results.} We provide an additional example in Appendix \ref{appendix: example} to show the limitation of existing decomposition policies.}
    \label{fig: framework and example}
\end{figure*}

\section{Introduction}
The \textit{Decompose-Then-Verify} paradigm has been widely used in fact-checking systems, as it reduces the claim complexity and makes the factuality evaluation fine-grained and easy \citep{min-etal-2023-factscore, chern2023factoolfactualitydetectiongenerative, 
chen2023felmbenchmarkingfactualityevaluation,
kamoi2023wicerealworldentailmentclaims,
wei2024longformfactualitylargelanguage, song-etal-2024-veriscore}. 
The paradigm comprises two components: (1) a decomposer, which leverages a large language model (LLM) guided by a decomposition policy\textemdash typically hand-crafted prompts\textemdash to break claims into subclaims,\footnote{Following \citet{jiang2024corerobustfactualprecision}, we use \textit{claim} to denote original sentence to be verified and \textit{subclaims} the result of decomposition.} and (2) a verifier, which similarly utilizes LLM paired with a verification policy (e.g., retrieving evidence to assist verification). \autoref{fig: framework and example} provides an illustration. 

In this work, we first systematically investigate how decomposition policies could affect verification through subclaim \textit{atomicity}\textemdash a metric we introduced for quantifying information density. We find that \textit{different verifiers achieve optimal verification confidence at distinct input atomicity}.
We define $\text{atomicity} = \text{log}_2(\text{\# atomic information})$, where one piece of atomic information is an utterance conveying a single nontrivial fact (e.g., \texttt{``Owen made a comeback in 2017''} in \autoref{fig: framework and example}). Higher atomicity means a subclaim is more coarse-grained and information-rich, which ostensibly implies lower verifier confidence, yet our finding indicates this is not always the case.

The above finding reveals that existing \textit{prompt-based decomposition policies do not always generate subclaims with optimal atomicity, resulting in suboptimal verification results}.
For instance, FActScore \citep{min-etal-2023-factscore} formulates its decomposition policy using eight annotated demonstrations to generate intended atomic subclaims. Our experiments find that verifiers, such as Inst-Llama-7B with the evidence retrieval verification policy shown in \autoref{fig: framework and example}, do not exhibit optimal performance at the atomicity level (i.e., $\text{atomicity}=0$) featured by the given decomposition policy.
Thus, there is a performance gap between the decomposer and verifier in terms of atomicity, which remains unaddressed. We further discuss the limitations of prior studies in the related work \S\ref{subsec: decompose then verify}.

To solve the above problem, we propose \textit{\method{}}\textemdash a novel framework to learn a decomposition policy tailored to the downstream verifier. Our approach is compatible with any existing fact-checking systems where both the decomposition and verification LLMs are given and frozen.
Unlike existing decomposition policies that use a single prompt call to generate subclaims \citep{min-etal-2023-factscore, wei2024longformfactualitylargelanguage, hu2024decompositiondilemmasdoesclaim, wanner2024closerlookclaimdecomposition}, 
we formulate \method{} as a \textit{Markov Decision Process} \citep{puterman2014markov} that involves a sequence of decomposition calls. At each step, the policy determines whether a subclaim should be decomposed, with the newly generated subclaims passed to the verifier that returns verification confidence change as a reward.\footnote{We provide a detailed justification of the reward design in \S\ref{subsec: implement dynamic decomposition policy}.}
Thus, this formulation enables the problem to be tackled using on-policy reinforcement learning (RL).

The learned policy, requiring only 4.73M parameters, significantly improves steerability over decomposition processes by dynamically decomposing claims to the verifier-perferred atomicity level.
Extensive experiments show that it outperforms baseline decomposition policies \citep{min-etal-2023-factscore, kamoi2023wicerealworldentailmentclaims, wanner2024closerlookclaimdecomposition}, improving both verification confidence (by 0.07) and accuracy (by 0.12) across varying verifiers, datasets, and atomicity levels.
In summary, our contributions are twofold:
\begin{itemize}[leftmargin=*]
    \item \textbf{Our study exposes the impact of subclaim atomicities on verifiers.} We find that each verifier prefers a distinct optimal input atomicity, yet existing decomposition policies hardly achieve the optimum.
    \item \textbf{We introduce a RL framework designed to bridge the performance gap between decomposers and verifiers.} It learns a decomposition policy to dynamically adjust the atomicity of subclaims tailored to downstream verifiers, thereby optimizing verification results.
\end{itemize}

\section{Methodology}
\label{sec: methodology}
We formulate finding the optimal decomposition policy for optimal verification as a bilevel optimization problem.
Specifically, given a decomposition LLM $\mathcal{D}$, a verification LLM $\mathcal{V}$, verification policy $\pi_v$, and a claim dataset $\{(C_i, Y_i)\}$ with binary factuality labels $Y_i$, we aim to find an optimal decomposition policy $\pi_d$ such that decomposed subclaims $\{c_j\}$ maximize the verification accuracy:
\begin{align}
    \label{eq: objective}
    &\max_{c\in\{c_j\},\pi_d}\mathbb{E}_i\Big[\mathbbm{1}\big(Y_i = \bigwedge_{c}\mathcal{V}(c\mid \pi_v)\big)\Big], \\ \nonumber
    &\text{subject to} \\\nonumber
    &\quad\pi_d \in \arg\max_{\pi_d}f(\{c_j\}, \pi_d),\;\{c_j\} \sim \mathcal{D}(C_i \mid \pi_d),
\end{align}
where verifier $\mathcal{V}(\cdot \mid \pi_v)$ returns the prediction label conditioned on its verification policy. We determine the claim is true if and only if all its subclaims are true, implemented via a logical AND operator $\bigwedge$. $f(,)$ represents the lower-level constraint, which observes the decisions $\{c_j\}$ made at the upper level and optimizes its own policy $\pi_d$ accordingly \citep{7942105}. Decomposer $\mathcal{D}$ is the upper-level constraint, ensuring that all subclaims are generated by it under a feasible decomposition policy that is lower-level optimal.

By formulating the problem as a bilevel optimization, we can simultaneously refine the upstream decomposition policy and optimize the downstream verification task, ensuring that the decomposition policy is aligned with the verifier to achieve optimal overall performance. 
Bilevel optimization is known to be strongly NP-hard \citep{linearbilevelprogramming}, and research has shown that it can be alternatively approximated using online stochastic approximation \citep{9435807}. Therefore, we propose our \method{} as an advantage actor-critic (A2C) style \citep{pmlr-v48-mniha16} RL solution to approximate Eq.\ref{eq: objective}. It enables the decomposition policy to learn directly from the verifier, dynamically converging toward an optimistic bilevel optimum.

\subsection{Overview of Dynamic Decomposition}
Unlike most of the prior work that applies its decomposition policy only once, we iteratively generate decomposition calls from the learned decomposition policy. The call is to either request the decomposition LLM to decompose the current subclaim or not. 
Specifically, given a temporary subclaim list $\{c_j\}$ and a target subclaim to be decomposed $c^\ast \in \{c_j\}$ sampled from it, we perform decomposition $\mathcal{D}(\cdot \mid \pi_d)$ in Eq.\ref{eq: objective} as:
\begin{align}
    \label{eq: decompose}
    \{c_j\}^\ast &=\begin{cases}
    \mathcal{D}(c^\ast), & \text{if $\pi_d(\{c_j\})=\text{decompose}$}\\
    \varnothing , & \text{if $\pi_d(\{c_j\})=\text{not decompose}$}
  \end{cases},\\
  \{c_j\} &\leftarrow \kappa(\{c_j\}, \{c_j\}^\ast).
  \label{eq: MDP state transition}
\end{align}
We repeat the above process until all subclaims in $\{c_j\}$ have been decided not to decompose further. 

Therefore, this can be formulated as a finite MDP defined as $M = (\mathcal{S}, \mathcal{A}, \kappa, r)$. $\mathcal{S}$ represents the state space, and $\mathcal{A}$ is the action space which includes two actions in our case: decompose or not to decompose. $\kappa: \mathcal{S} \times \mathcal{S}^\ast\rightarrow \mathcal{S}$ is the state transition function that replaces the target subclaim $c^\ast$ in the subclaim list $\{c_j\}$ with its decomposition results $\{c_j\}^\ast$. $r: \mathcal{S} \times \mathcal{S}^\ast\rightarrow \mathcal{A}$ is the immediate reward received from the verifier after state transition.

\subsection{Implement Dynamic Decomposition Policy}
\label{subsec: implement dynamic decomposition policy}
In this section, we elaborate on how we implement our \method{} based on the MDP formulation proposed above. 

\paragraph{Atomicity state:} To find the optimal atomicity that is favored by the verifier, we create an \textit{atomicity state} reflecting the overall atomicity of current subclaims $\{c_j\}$ at step $t$. Each atomicity state is a $d$ dimension vector $s_t \in \mathbb{R}^d$.

\paragraph{State transition:} Similar to the work of \citet{chen-etal-2024-learning-retrieve}, we use a trainable Gated Recurrent Unit (GRU) \citep{cho-etal-2014-learning} to model state transition function $\kappa$ in Eq.\ref{eq: MDP state transition}:
\begin{align}
    s_{t+1} = \text{GRU}\big[s_t, (1 + \sigma(\Delta\text{Info}))\text{Enc}(\{c_j\})\big],
    \label{eq: state transition}
\end{align}
where $\text{Enc}(\cdot): T\rightarrow \mathbb{R}^d$ is a textual encoder that maps the text sequence to a $d$-dimensional embedding and $\sigma: \mathbb{R}\rightarrow \mathbb{R}$ is the sigmoid function. We compute $\Delta\text{Info}$ as the average Conditional Pairwise Mutual
Information (CPMI) \citep{jiang2024corerobustfactualprecision} difference between the target subclaim $c^\ast$ and its decomposed results $\{c_j\}^\ast$, which basically quantifies how much information is lost from each subclaim after decomposition:
\begin{align}
    \Delta\text{Info} = \mathbb{E}_{c\in\{c_j\}^\ast} \Big[ \log\frac{P(c\mid \mathcal{H})}{P(c^\ast\mid \mathcal{H})}\Big],
    \label{eq: information loss}
\end{align}
where $P(\cdot \mid \mathcal{H})$ measures the entailment probability between a claim and a pre-defined set of tautological and bleached claims $\mathcal{H}$ (e.g., ``{\fontfamily{qcr}\selectfont \{topic\}} is a person'', ``{\fontfamily{qcr}\selectfont \{topic\}} exists''). In practice, this conditional probability is estimated as $\max_{h\in\mathcal{H}}P(\cdot \mid h)$ for better stability \citep{jiang2024corerobustfactualprecision}.
As both $c^\ast$ and its subclaims $c \in \{c_j\}^\ast$ getting syntactically closer to bleached claims through iterative decomposition, $\Delta\text{Info}$ tend to decrease because of the diminishing marginal information loss (e.g., $\frac{P(\texttt{``Kruger was a religious man''}\mid \mathcal{H})}{P(\texttt{``Kruger was a deeply ... faith''} \mid \mathcal{H})} \gg \frac{P(\texttt{``Kruger was a man''}\mid \mathcal{H})}{P(\texttt{``Kruger was a religious''} \mid \mathcal{H})}$; see \autoref{fig: avg info loss} in Appendix \ref{appendix: experiment results} for experimental justification.).

Therefore, we design $\Delta\text{Info}$ as a metric to quantify localized atomicity change caused by the current decomposition call, and $\text{Enc}(\{c_j\})$ reflects the overall atomicity of all subclaims in terms of semantics. Multiplying global decomposition embeddings by the local atomicity change in Eq.\ref{eq: state transition} preserves original semantics \citep{NIPS2017_3f5ee243} while revealing hidden states regarding atomicity of current subclaims.

\paragraph{Action:} We define $\mathcal{A}$ as having only two actions: 1 (decompose) or 0 (not decompose). For each atomicity state, the action is sampled from a policy distribution, $a_t \sim \pi_d(a_t\mid s_t)$, namely our decomposition policy. $\pi_d$ is trained to determine when to decompose a claim until achieving the desired atomicity state that maximizes the reward.

\begin{figure}[ht]
    \setlength{\belowcaptionskip}{-10pt}
    \centering
    \includegraphics[width=0.95\linewidth]{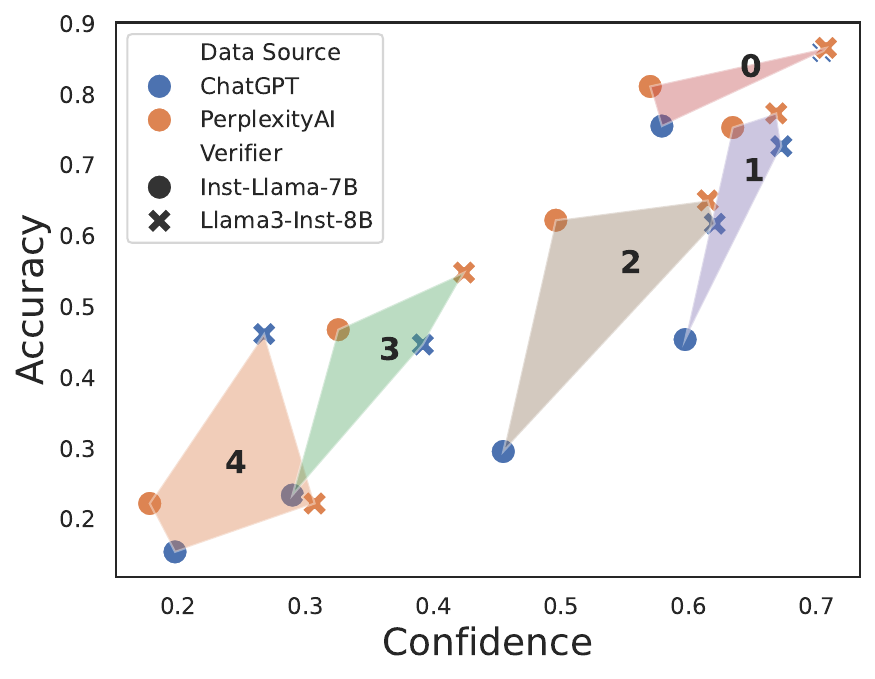}
    \caption{Verification confidence versus accuracy. The number in each convex hull denotes the claim atomicity. 
    \textbf{Irrespective of data sources, atomicities, and verifiers, verification confidence exhibits a strong positive correlation with accuracy (0.88 Pearson's r).}
    }
    \label{fig: confidence versus accuracy}
\end{figure}

\paragraph{Reward:} Due to the lack of ground-truth labels for newly generated subclaims, accuracy-based evaluation is not feasible for evaluating verification improvement after decomposition. To address this limitation, we introduce verification confidence, a label-free proxy for accuracy that is computable for all subclaims. It is defined as the absolute probability difference between positive and negative verification labels: 
\begin{align}
\label{eq: verification confidence}
\text{Conf}(c, \mathcal{V}, \pi_v) =  \big|P_{\mathcal{V}}(\texttt{True}& \mid c, \pi_v) - \\ \nonumber 
&P_{\mathcal{V}}(\texttt{False} \mid c, \pi_v)\big|.
\end{align} 
Verification confidence measures \textit{how much certainty a verifier has in making a verification}, which we find to strongly correlate with verification accuracy in \autoref{fig: confidence versus accuracy}.

Our reward is the verification confidence change before and after decomposition. This design encourages the policy to perform decomposition when the verifier is more confident in evaluating the factuality of generated subclaims:
\begin{align}
    r_t = \underbrace{\mathbb{E}_{c\in\{c_j\}^\ast}\Big[\text{Conf}(c, \mathcal{V}, \pi_v)\Big]}_\text{After Decomposition} - \underbrace{\text{Conf}(c^\ast, \mathcal{V}, \pi_v)}_\text{Before Decomposition}.
    \label{eq: reward}
\end{align}

\begin{figure}[ht]
    \centering
    \includegraphics[width=\linewidth]{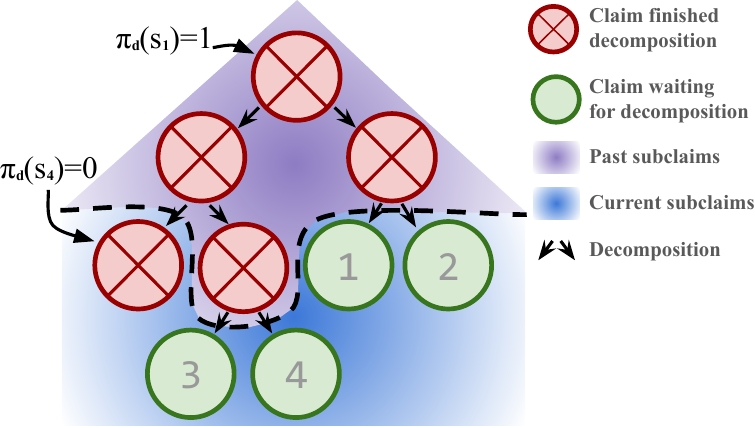}
    \caption{Breadth-first order sampling for dynamic decomposition. We perform binary decomposition for each claim. The number in the node represents its sampling priority in the decomposition process. We first sample out subclaims at the same atomicity level, with newly generated subclaims queued in a FIFO (first-in-first-out) order.}
    \label{fig: bfs sampling}
\end{figure}

\paragraph{Breadth-First Order Decomposition:} In \method{}, each decomposition call could generate a new list of subclaims at a lower level of atomicity, which are then queued for further decomposition. Therefore, the order of decomposition becomes matter. A depth-first approach, where a single subclaim is continually decomposed until reaching the lowest possible atomicity level, can result in significant variance in atomicity among all subclaims, which in turn leads to high variance in modeling the state (Eq.\ref{eq: state transition}). Hence, we employ a breadth-first strategy to prioritize the decomposition of subclaims at higher atomicity. We provide an illustration in \autoref{fig: bfs sampling}.

\subsection{Train Dynamic Decomposition Policy}
We employ PPO \citep{schulman2017proximalpolicyoptimizationalgorithms} in A2C style to train our dynamic decomposition policy given its effectiveness and stability \citep{engstrom2020implementationmattersdeeppolicy}. We model policy function $\pi_d: \mathbb{R}^d \rightarrow \mathbb{R}^2$ as an MLP (Multi-Layer Perceptron) that outputs a two-dimensional normalized vector. A core component of the PPO objective function is the clipped surrogate term, which constrains the policy change during the optimization process. Given a finite decomposition trajectory $\{(a_1,s_1), (a_2, s_2), \cdots, (a_T,s_T)\}$, we have
\begin{align}
    L^{\text{clip}} = \mathbb{E}_t\Big[\min\big(\rho_t, \text{clip}(\rho_t, 1-\epsilon, 1+\epsilon)\big)\hat{A}_t\Big],
    \label{eq: clipped surrogate term}
\end{align}
where $\rho_t = \frac{\pi_d(a_t\mid s_t)}{\pi_d^{\text{old}}(a_t\mid s_t)}$ is a probability ratio to estimate the divergence between old and current policy. The hyperparameter $\epsilon$ sets the clipping boundary for $\rho$ to fall between $[1-\epsilon, 1+ \epsilon]$. $\hat{A}_t$ is the advantage at step $t$ which measures how better taking the action $a_t$ at state $s_t$ is compared to the average value of the state. 

To calculate the average value of a state, we create another trainable MLP as a value function, $v: \mathbb{R}^d \rightarrow \mathbb{R}$, to map the state to its corresponding value. Then, we estimate the advantage $\hat{A}_t$ using GAE (Generalized Advantage Estimator; \citet{schulman2018highdimensionalcontinuouscontrolusing}):
\begin{align}
    &\hat{A}_t = \delta_t + (\gamma\lambda)\delta_{t+1} + \cdots +(\gamma\lambda)^{T-t+1}\delta_{T-1}, \nonumber \\
    &\text{where } \delta_t = r_t + \gamma v(s_{t+1}) - v(s_t).
    \label{eq: gae}
\end{align}
$r_t$ is the reward defined in Eq.\ref{eq: reward}. $\delta_t$ is the TD residual of value function with discount factor $\gamma$ \citep{rl_intro}. Another hyperparameter $\lambda$ controls the trade-off between bias and variance in advantage estimation. We use the squared-error loss to train our value function: $(v(s_t) - v^{\text{target}}_t)^2 \approx \big(v(s_t) - (\hat{A}_t + v(s_t))\big)^2 = \hat{A}_t^2$. Therefore, our final PPO objective function with entropy bonus term $S[\pi_d](s_t)$ becomes:
\begin{align}
    L^{\text{PPO}} = \mathbb{E}_t\Big[L^{\text{clip}} - c_1 \hat{A}_t^2 + c_2S[\pi_d](s_t)\Big],
    \label{eq: ppo objective}
\end{align}
where $c_1$ and $c_2$ are coefficients. We perform gradient descent on $-L^{	\text{PPO}}$ to maximize the above objective function. \autoref{alg: train dynamic decomposition} outlines the procedure.

\begin{algorithm}[ht]
\small
\caption{Train Dynamic Decomposition Policy}
\begin{flushleft}
    \textbf{Input:} decomposition LLM $\mathcal{D}$, verification LLM $\mathcal{V}$, decomposition policy $\pi_d$, verification policy $\pi_v$, value function $v$, state transition model GRU, initial atomicity state $s_0$, claims $\{C_i\}$ to be verified.
\end{flushleft}
\begin{algorithmic}[1]
\While{not done}
\Statex \;\;\; \textcolor{darkgreen}{\# update replay buffer}
    \For{step = $1,\cdots,m$}
    \State $C\sim\{C_i\}, \{c_j\} = C, s_t = s_0$ \Comment{\textcolor{darkgreen}
    {initialization}}
        \While{not finish decomposition}
        \State $s_{t} \leftarrow \text{start atomicity state}$
        \State $c^\ast \xleftarrow[]{\text{BF Sampling}} \{c_j\}$ \Comment{\textcolor{darkgreen}{get target subclaim}}  
        \State $a_{t} \leftarrow \text{sample action from } \pi_d(a_t\mid s_t)$
        \State LLM $\mathcal{D}$ decompose $c^\ast$ following Eq.\ref{eq: decompose} 
        \State $\{c_j\} \leftarrow \text{update subclaim list following Eq.\ref{eq: MDP state transition}}$
        \State $r_t \leftarrow \text{reward from Eq.\ref{eq: reward}}$
        \State $s_{t+1} \leftarrow \text{end atomicity state updated by Eq.\ref{eq: state transition}}$
        \State Record $(a_t, r_t, s_t, s_{t+1})$ into replay buffer $R$
        \EndWhile
    \EndFor

\Statex \;\;\; \textcolor{darkgreen}{\# train \method{} policy}
\State $\pi_d^{\text{old}} \leftarrow \pi_d$
\For{each update step}
    \State Sample minibatch $M$ from replay buffer $R$
    \For{each $(a_t, r_t, s_t, s_{t+1})$ in $M$}
        \State $\hat{A}_t \leftarrow \text{advantage from Eq.\ref{eq: gae}}$
        \State $\rho_t \leftarrow \frac{\pi_d(a_t\mid s_t)}{\pi_d^{\text{old}}(a_t\mid s_t)}$
    \EndFor
    \State $L^{\text{clip}} \leftarrow \text{clipped surrogate term from Eq.\ref{eq: clipped surrogate term}}$
    \State $L^{\text{PPO}} \leftarrow \text{objective function from Eq.\ref{eq: ppo objective}}$ 
    \State Update $\pi_d, v, \text{GRU}$ using $-L^{\text{PPO}}$ through GD
\EndFor
\EndWhile
\end{algorithmic}
      \label{alg: train dynamic decomposition}
\end{algorithm}

\section{Experiment Setup}
\label{sec: experiment setup}

\subsection{Dataset Construction}
\label{subsec: dataset construction}
To evaluate the effect of decomposition policies on verification, we construct two claim datasets from FActScore \citep{min-etal-2023-factscore}, whose original claims are sourced from \textbf{ChatGPT} \citep{openai2022chatgpt} and \textbf{PerplexityAI} \citep{PerplexityAI2023}.\footnote{For simplicity, we refer to each dataset by the name of its source LLM.} Each dataset contains claims with 6 different atomicities ranging from -1 to 4.
Specifically, we consider the given human-annotated atomic subclaims as the base with atomicity 0, because they typically are self-contained sentences with single factual information (e.g., \texttt{``The String is a collection of poetry''}). 
We construct higher-level subclaims by recursively merging pairs at the same atomicity level from the base up until atomicity 4, which is the highest level that cannot be merged further.
We also decompose the base subclaims to a lower level at atomicity -1, where subclaims are partially tautological (e.g., \texttt{``String exists''}, \texttt{``String is a collection''}, \texttt{``The collection is composed of poetry''}).\footnote{We use \texttt{gpt-3.5-turbo-0125} to perform the decomposition and name the decomposition policy as FActScore-Atom whose prompt can be found in Appendix \ref{appendix: prompts}.}
Thus, each subclaim is purposefully built to contain $2^{\text{atomicity}}$ pieces of information.
We provide data statistics in Appendix \ref{appendix: dataset statistics}.

\subsection{Models}
\paragraph{Decomposers}
Because decomposition requires a deep understanding of both the semantic and syntactic aspects of the given claims, we use the following two open-source and widely-recognized models as decomposition LLM: \texttt{Llama3-Inst-70B} \citep{grattafiori2024llama3herdmodels} and \texttt{DeepSeek-V3} \citep{deepseekai2024deepseekv3technicalreport}.

\begin{figure*}[t]
    \setlength{\belowcaptionskip}{-10pt}
    \setlength{\belowcaptionskip}{-10pt}
    \centering
    \includegraphics[width=\linewidth]{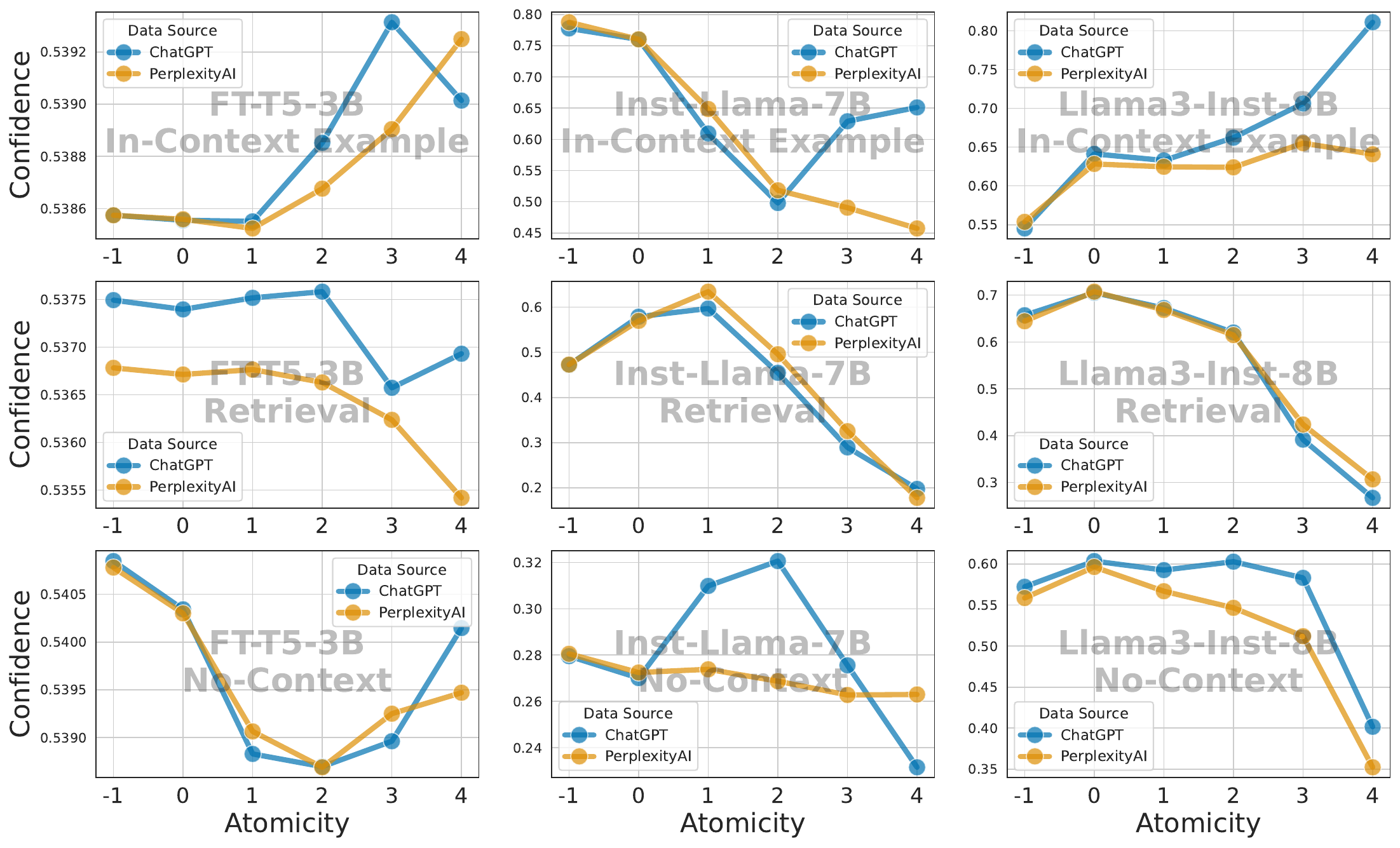}
    \caption{Verification confidence across atomicities. \textbf{Evidently, each verifier has its own preferred input atomicity at which the verification confidence peaks}. Even when utilizing the same verification policy, such as retrieval, different verifiers exhibit distinct preferences, and vice versa.}
    \label{fig: granularity preference}
\end{figure*}

\paragraph{Verfiers}
Following \citet{min-etal-2023-factscore}, we determine factual labels by comparing the conditional probability of \texttt{True} and \texttt{False} from the verifier. We experiment the following three verification LLMs: a T5-3B \citep{raffel2023exploringlimitstransferlearning} fine-tuned on FActScore for the factuality verification task (\texttt{FT-T5-3B}), a Llama-7B \citep{touvron2023llamaopenefficientfoundation} trained on Super Natural Instructions \citep{wang-etal-2022-super} (\texttt{Inst-Llama-7B}), and a pretrained and instruction tuned Llama3-8B model \citep{grattafiori2024llama3herdmodels} (\texttt{Llama3-Inst-8B}). We employ three verification policies, each utilizing differently constructed prompts to assist verification: (1) \textbf{Retrieval} retrieves relevant passages from a database as evidence; (2) \textbf{In-Context Example} provides verification demonstrations to instruct verification;\footnote{In-context examples can be found in \autoref{appendix: prompts}.} and (3) \textbf{No-Context} directly asks the verifier for predictions, as inspired by \citet{kadavath2022languagemodelsmostlyknow}.

\subsection{Training}
\paragraph{Initialization} We model each policy and value function as a two-layer fully connected perceptron with a ReLU activation function \citep{agarap2019deeplearningusingrectified}.
The total number of trainable parameters is $4.73$M. We perform binary decomposition on each subclaim during our \method{}, aligning with the definition of atomicity (using a logarithm with base 2), to ensure maximal exploration of the subclaim space. Please refer to Appendix \ref{appendix: training initialization} for more details.

\paragraph{Data} We train our policy on two constructed claim datasets across atomicity $[1, 4]$.

\paragraph{Hyperparameters} Please see Appendix \ref{appendix: hyperparameters}.

\subsection{Baselines and Metrics}
We compare our trained decomposition policy (hereafter denoted as \textbf{\short}) to existing decomposition policies, including \textbf{FActScore} \citep{min-etal-2023-factscore}, \textbf{WICE} \citep{kamoi2023wicerealworldentailmentclaims} and \textbf{R-ND} \citep{wanner2024closerlookclaimdecomposition}, on both \textbf{verification confidence} (Eq.\ref{eq: verification confidence}) and \textbf{accuracy} (Eq.\ref{eq: objective}). These works typically apply heuristic splitting prior to neural decomposition, thus their decomposition policies are primarily designed for subclaims under atomicity 2. To ensure a fair comparison, our evaluation is conducted on claims on atomicity within $[0,2]$. We also evaluate \short{} against a modified version of the FActScore policy, \textbf{FActScore-Atom}, which is designed to decompose human-annotated atomic subclaims (atomicity 0) to partially trivial subclaims (atomicity -1) in \S\ref{subsec: dataset construction}.

\setlength{\aboverulesep}{1.4pt}
\setlength{\belowrulesep}{1.4pt}
\begin{table*}[t]
\setlength{\belowcaptionskip}{-10pt}
\centering
\small
\begin{tabular}{@{}>{\centering\arraybackslash}p{1.2cm}|p{2.7cm}ccccc@{}}
\toprule
\multirow{2}{*}{\textbf{Atomicity}} &
\multirow{2}{*}{\begin{tabular}[c]{@{}l@{}}
  \textbf{Decompose Policy} $\rightarrow$\\
  \textbf{Verify Policy}$\downarrow$
\end{tabular}} & FActScore & FActScore-Atom & WICE & R-ND & \textbf{\short} \\
& & \multicolumn{5}{c}{\textbf{\textit{Verification Confidence [0-1] $\uparrow$ / Verification Accuracy [0-1] $\uparrow$}}} \\
\midrule
\multirow{8}{*}{0} 
  & \multicolumn{6}{l}{\it Data Source: ChatGPT} \\
  & Retrieval          & 0.627 / 0.666 & 0.618 / 0.796 & 0.431 / 0.782 & 0.556 / 0.449 & 0.600 / 0.789 \\
  & In-Context Example & 0.677 / 0.454 & 0.677 / 0.388 & 0.724 / 0.451 & 0.714 / 0.401 & 0.715 / 0.428 \\
  & No-Context         & 0.557 / 0.457 & 0.526 / 0.566 & 0.374 / 0.525 & 0.527 / 0.378 & 0.547 / 0.551 \\
\cmidrule(lr){2-7}
  & \multicolumn{6}{l}{\it Data Source: PerplexityAI} \\
  & Retrieval          & 0.629 / 0.559 & 0.611 / 0.755 & 0.435 / 0.762 & 0.541 / 0.241 & 0.612 / 0.799\\
  & In-Context Example & 0.681 / 0.266 & 0.670 / 0.197 & 0.726 / 0.308 & 0.711 / 0.172 & \textbf{0.733} / 0.301\\
  & No-Context         & 0.555 / 0.352 & 0.515 / 0.471 & 0.380 / 0.475 & 0.519 / 0.166 & 0.542 / \textbf{0.554} \\
\midrule
\multirow{8}{*}{1} 
  & \multicolumn{6}{l}{\it Data Source: ChatGPT} \\
  & Retrieval          & 0.609 / 0.739 & 0.611 / 0.815 & 0.527 / 0.755 & 0.541 / 0.635 & \textbf{0.654} / 0.758 \\
  & In-Context Example & 0.714 / 0.635 & 0.705 / 0.627 & 0.658 / 0.631 & 0.749 / 0.610 & \textbf{0.809} / 0.609 \\
  & No-Context         & 0.549 / 0.619  & 0.521 / 0.550 & 0.437 / 0.606 & 0.508 / 0.610 & \textbf{0.567} / 0.512 \\
\cmidrule(lr){2-7}
  & \multicolumn{6}{l}{\it Data Source: PerplexityAI} \\
  & Retrieval          & 0.610 / 0.493 & 0.615 / 0.68 & 0.527 / 0.597 & 0.515 / 0.232 & \textbf{0.651} / \textbf{0.753} \\
  & In-Context Example & 0.721 / 0.247 & 0.711 / 0.253 & 0.684 / 0.27 & 0.746 / 0.22 & \textbf{0.791} / \textbf{0.347} \\
  & No-Context         & 0.535 / 0.260 & 0.506 / 0.350 & 0.423 / 0.273 & 0.506 / 0.223 & \textbf{0.559} / \textbf{0.437} \\
\midrule
\multirow{8}{*}{2} 
  & \multicolumn{6}{l}{\it Data Source: ChatGPT} \\
  & Retrieval          & 0.616 / 0.844 & 0.639 / 0.887 & 0.588 / 0.809 & 0.547 / 0.852 & \textbf{0.644} / 0.652 \\
  & In-Context Example & 0.731 / 0.835 & 0.725 / 0.835 & 0.665 / 0.835 & 0.744 / 0.835 & \textbf{0.824} / 0.750 \\
  & No-Context         & 0.553 / 0.809 & 0.545 / 0.800 & 0.507 / 0.835 & 0.515 / 0.835 & \textbf{0.583} / 0.509 \\
\cmidrule(lr){2-7}
  & \multicolumn{6}{l}{\it Data Source: PerplexityAI} \\
  & Retrieval          & 0.622 / 0.483 & 0.639 / 0.601 & 0.592 / 0.546 & 0.529 / 0.406 & 0.633 / \textbf{0.664} \\
  & In-Context Example & 0.755 / 0.392 & 0.734 / 0.378 & 0.675 / 0.392  & 0.751 / 0.392 & \textbf{0.823} / \textbf{0.464} \\
  & No-Context         & 0.543 / 0.378 & 0.535 / 0.378 & 0.492 / 0.392 & 0.509 / 0.392 & \textbf{0.544} / \textbf{0.421} \\
\bottomrule
\end{tabular}
\caption{Comparison of our \short{} over baselines on the test dataset. Each metric is scaled from 0 to 1. $\uparrow$ indicates higher values are preferred. We employ decomposition LLM \texttt{Llama3-Inst-70B} and verification LLM \texttt{Llama3-Inst-8B}.
\textbf{\short{} consistently outperforms on atomicity 1 and 2, achieving an average improvement of 0.07 in verification confidence across two datasets and three verifiers, and a 0.12 average improvement in verification accuracy for claims sourced from PerplexityAI.}
}
\label{table: evaluation results}
\end{table*}

\section{Results and Analysis}
\label{sec: results and analysis}
We first study the effect of subclaim atomicity on verification (\S\ref{subsec: effect of atomicity}), followed by evaluating \method{} against existing baselines (\S\ref{subsec: results on dynamic decomposition}) and ablation study (\S\ref{subsec: ablation study}).

\subsection{Effect of Subclaim Atomicity on Verification}
\label{subsec: effect of atomicity}
Prior works have found that decomposition policy influences verification results and does not guarantee consistent verification improvement across varying input length and verifier strength \citep{jiang2024corerobustfactualprecision, wanner2024closerlookclaimdecomposition, hu2024decompositiondilemmasdoesclaim}. In this study, we take one step further by looking into the following two questions: \textit{How can we quantify the impact of decomposition policy on verification, and why does this influence not yield consistent improvements across different verifiers?}

\paragraph{Why use verification confidence?} Our experiment finds a strong correlation between verification confidence and accuracy across various conditions (\autoref{fig: confidence versus accuracy}), including different datasets, atomiticity levels, and verifies. Additionally, verficaition confidence is more accessible than accuracy as it does not require ground-truth labels. These properties support verification confidence as \textit{a reliable metric for evaluating the impact of decomposition policies} and as an ideal signal funneling back to policy.

\paragraph{Each verifier has its own atomicity optimum.} 
We investigate how atomicity, a key characteristic of decomposed subclaims, affects verification. \autoref{fig: granularity preference} shows verification confidence change across different atomicity levels. We find that each verifier, which is a verification LLM with a specific verification policy, achieves peak verification confidence at a distinct optimal atomicity. For instance, Llama3-Inst-8B with a retrieval verification policy consistently performs best at atomicity 0 on both the ChatGPT and PerplexityAI datasets, whereas FT-T5-3B with retrieval reaches its optimum at atomicity 2. This observation answers the second question that \textit{the different preference for atomicity makes existing static decomposition policies hard to find optimal subclaims that bring consistent verification improvement.}

\subsection{Dynamic Decomposition Results}
\label{subsec: results on dynamic decomposition}
\autoref{table: evaluation results} shows the evaluation results of decomposer \texttt{Llama3-Inst-70B} with \short{} policy on various verifiers, datasets, and claim atomicities. Evidently, \short{} consistently achieves the highest verification confidence for claims with atomicity 1 and 2, outperforming the four baselines by an average margin of 0.07.
This result aligns with our findings from \autoref{fig: granularity preference}, where the verification LLM for \autoref{table: evaluation results} (i.e., \texttt{Llama3-Inst-8B}) generally achieves the best performance at atomicity 0. 
In other words, an effective decomposition policy should strategically decompose claims to lower atomicities, ideally to 0, which is an ability our \short{} excels. Conversely, since claims are already near optimal, \short{} does not always achieve higher verification confidence than other baselines when $\text{atomicity} = 0$.

We also evaluate \short{} on verification accuracy and observe a notable average improvement of 0.12 for claims from PerplexityAI with atomicity 1 and 2. We repeat the experiments with another decomposition LLM, \texttt{DeepSeek-V3}, and observe similar improvement in verification accuracy (see \autoref{table: comparison results ds-v3} in Appendix \ref{appendix: experiment results}). 

\subsection{Case Study}
A critical question emerges when adopting verification confidence as a proxy for accuracy: \textit{whether the policy would over-optimize for confidence in ways that did not actually improve accuracy?} Our findings indicate that this phenomenon does happen, primarily due to inherent limitations in verifier capabilities. As shown in \autoref{table: evaluation results}, on the PerplexityAI dataset, we observe aligned improvements in both verification confidence and actual accuracy. However, the ChatGPT dataset demonstrates substantial gains in verifier confidence with only negligible improvements in accuracy.

We posit that the disparity stems from fundamental differences in verifier capability relative to data complexity. Specifically, we find claims in the ChatGPT dataset are generally more complex and less grounded than those in the PerplexityAI dataset, making them inherently more difficult to verify. In contrast, PerplexityAI claims are typically more concise and feature explicit reference through indexed citations, enabling even less capable verifiers (\texttt{Llama3-Inst-8B} in this case) to evaluate factuality with greater confidence and accuracy. We provide examples in Appendix \ref{appendix: data example} to illustrate this difference.

Our observations indicate that verifier capability is a pivotal factor in the success of \method{}, exemplifying a ``bucket effect'' where the weakest component constrains system performance. When paired with a sufficiently strong verifier, the \method {} can optimize the entire system in tandem, ensuring that confidence improvements translate into genuine accuracy gains.

\subsection{Ablation Study}
\label{subsec: ablation study}
\autoref{table: ablation} display our ablation study on different components of \short{} regarding \textit{algorithm design} and \textit{training data selection}. We use the decomposition LLM \texttt{Llama3-Inst-70B}, and the verifier \texttt{Llama3-8B-Inst} with a retrieval verification policy. The ablation experiments are conducted on claims with atomicity 4.

\setlength{\aboverulesep}{.4ex}
\setlength{\belowrulesep}{.65ex}
\begin{table}[ht]
    \centering \small
    \setlength{\tabcolsep}{4pt}
    \begin{tabular}{lc}
    \toprule
        \bf Variants & \bf Verification Confidence  \\
    \midrule
        \textbf{\short} & \textbf{0.446}  \\ 
        \;\;\textit{$-$ one layer NN} & 0.398 (-0.048) \\
        \;\;\textit{$-$ binary; $+$ triple decompose} & 0.424 (-0.022)  \\ 
        \;\;\textit{$-$ entropy bonus} & 0.356 (-0.090) \\ \midrule
        \;\;\textit{$-$ data on atomcitity 1} & 0.353 (-0.093) \\ 
        \;\;\textit{$-$ data on atomicity 1,  2} & 0.356 (-0.090)  \\
        \;\;\textit{$-$ data on atomicity 1,  2, 3} & 0.401 (-0.045)  \\
    \bottomrule
    \end{tabular}
    \caption{Ablation study results of \short. \textit{``$-$'' indicates the removal of a component from} \short. For instance, ``$-$ one layer NN'' means modeling the policy and value functions using a single-layer perceptron instead of two layers. ``$-$ data on atomicity 1'' removes claims with atomicity 1 from the training data.}
    \label{table: ablation}
\end{table}

\paragraph{Decomposition exploration is effective for long-form text verification.} In experiment setup \S\ref{sec: experiment setup}, we model each policy and value function as a two-layer perceptron. We investigate whether a simple, shallow model can capture the atomicity state and find that reducing the network to a single layer results in lower verification confidence for decomposed subclaims. Similarly, switching from binary to triple decomposition at each step, which reduces the exploration of subclaims on different atomicities, also leads to declined verification confidence.
Furthermore, removing the entropy bonus term, which promotes action exploration, from the objective function leads to a substantial drop in verification confidence (-0.090). These findings demonstrate that \short{} benefits from diverse decomposition trajectories during training, which facilitate the search for subclaims with optimal atomicity.

\paragraph{Cross-atomicity training data stabilizes performance.} In \autoref{table: ablation}, we train \short{} using claims with atomicity ranging from 0 to 4 and evaluate on atomicity 4. 
We find that gradually removing claims of lower atomicity (from 1 to 3) from the training set negatively impacts verification performance, highlighting the importance of cross-atomicity data for improving the generalizability of \short{}. 
However, this negative effect diminishes as more data with irrelevant atomicities is removed ($-0.093 \rightarrow -0.045$). This suggests that increased exposure to claims of a specific atomicity during PPO rollouts enhances learning for that atomicity but cannot fully compensate for performance losses due to reduced atomicity coverage in training data.

\section{Related Works}
\label{sec: related works}

\subsection{Decompose-Then-Verify Paradigms}
\label{subsec: decompose then verify}
Unlike traditional fact-checking systems that focus on short and simple claims \citep{thorne-etal-2018-fever, sathe-etal-2020-automated, schuster-etal-2021-get, Chen_Bao_Sun_Zhang_Chen_Zhou_Xiao_Li_2022, guo-etal-2022-survey}, \textit{Decompose-Then-Verify} now becomes a typical approach in long-form text evaluation works, as it allows for more precise error identification and enhances the accuracy of verification by decomposing claims into shorter subclaims, which can then be independently validated \citep{min-etal-2023-factscore, chern2023factoolfactualitydetectiongenerative, kamoi2023wicerealworldentailmentclaims,
chen2023felmbenchmarkingfactualityevaluation, iqbal-etal-2024-openfactcheck, song-etal-2024-veriscore, chen-etal-2024-complex, wang2024factcheckbenchfinegrainedevaluationbenchmark}. However, how the decomposition and verification should be conducted is always underspecified.

\paragraph{Decomposition policies.} Existing factuality evaluation works have proposed various decomposition prompts revealing different characteristics of textual decomposition regarding precision \citep{min-etal-2023-factscore}, verifiability \citep{song-etal-2024-veriscore}, coverage \citep{wanner2024closerlookclaimdecomposition}, and atomicity \citep{stacey-etal-2024-atomic}. More recently, research has found that superficially fine-grained subclaims with trivial information could easily inflate verification precision \citep{jiang2024corerobustfactualprecision}, and decomposition benefits weaker verifiers more than stronger verifiers by generating simpler subclaims \citep{hu2024decompositiondilemmasdoesclaim}. Our  \method{} effectively addresses the over-optimization problem by providing necessary controllability over the decomposition process.

Moreover, static decomposition policies may struggle to handle input claims with varying fact density and often produce atomically homogeneous subclaims. We elaborate on these limitations in Appendix \ref{appendix: limitations of decomposition policies}.

\paragraph{Verification policies.} Policies for verification can be categorized according to the methods verifiers utilize to process subclaims and predict final labels. Popular processing methods include retrieving relevant evidence  \citep{kamoi2023wicerealworldentailmentclaims, wei2024longformfactualitylargelanguage}, constructing in-context exemplars \citep{kamoi2023wicerealworldentailmentclaims, song-etal-2024-veriscore}, generating claim-focused summarization \citep{chen-etal-2024-complex}, and simply zero-shot prompting \citep{kadavath2022languagemodelsmostlyknow, min-etal-2023-factscore}. 
The final label can be predicted by either gradient-based approaches (e.g., comparing logits of factual labels) \citep{chen-etal-2024-complex, tang-etal-2024-minicheck, milbauer-etal-2023-newssense, kamoi2023wicerealworldentailmentclaims, min-etal-2023-factscore} or searching for keywords (e.g., \texttt{True} or \texttt{False}) \citep{min-etal-2023-factscore, li-etal-2024-self, song-etal-2024-veriscore}. 

\subsection{RL in NLP Problems}
\label{subsec: rl in nlp}
Prior works have validated the use of RL in optimizing singular-task systems, such as identifying optimal exemplars for in-context learning \citep{zhang-etal-2022-active, lu2023dynamic, chen-etal-2024-learning-retrieve}. 
However, how RL can be applied to dual-task systems where multiple LLMs are involved remains underexplored. One concurrent work uses two LLMs as process reward and policy models, employing PPO to train them jointly for challenging reasoning tasks \citep{cui2025processreinforcementimplicitrewards}. 
In contrast, our \method{} first explores RL to solve a bilevel optimization problem in another dual-task system, \textit{Decompose-Then-Verify}, to reveal the nuanced characteristics of hierarchical LLM systems through their interactions.

\section{Conclusion}
We find that each verifier has an optimal atomicity where its verification confidence peaks. To leverage it, we introduce \method{} that optimizes claim verification by decomposing claims into verifier-preferred atomicity learned via on-policy optimization. Our policy stands out for its adaptability to diverse verifiers and input claim atomicities, outperforming existing baselines while adding only 4.73M parameters.

\section*{Limitations}
\paragraph{Different characteristics of decomposition.} While \method{} addresses the problem of misalignment between decomposer and verifier, we focus on the aspect of information density (i.e., atomicity). Although well-structured, self-contained, and verifiable subclaims could further improve verification, these aspects are beyond the scope of this paper. Future research could investigate other key characteristics of decomposition and explore how to amplify their positive effects on verification through \method{} (e.g., using a more powerful decomposer).

\paragraph{Evaluation metrics.} Our reward design relies on verification confidence rather than accuracy. We leave it for future works to acquire more ground-truth labels to effectively employ verification accuracy as feedback.

\section*{Acknowledgments}
This work was supported by NSF IIS-2119531, IIS-2137396, IIS-2142827, IIS-2234058, CCF-1901059, and ONR N00014-22-1-2507. The authors would like to thank Chihiro Taguchi, Katsumi Ibaraki and Demetrius Hernandez for their helpful input on earlier versions of this work. GPU machines for conducting experiments were provided by CRC cluster (\href{https://crc.nd.edu/}{https://crc.nd.edu/}).


\bibliography{custom}

\providecommand{\CNFX}[1]{{\em{\textrm{(#1)}}}}
\begin{thebibliography}{49}
\providecommand{\natexlab}[1]{#1}

\bibitem[{Agarap(2019)}]{agarap2019deeplearningusingrectified}
Abien~Fred Agarap. 2019.
\newblock \href {https://arxiv.org/abs/1803.08375} {Deep learning using rectified linear units (relu)}.
\newblock \emph{Preprint}, arXiv:1803.08375.

\bibitem[{AI(2023)}]{PerplexityAI2023}
Perplexity AI. 2023.
\newblock Perplexity.ai.
\newblock \url{https://www.perplexity.ai/}.

\bibitem[{Chen et~al.(2022)Chen, Bao, Sun, Zhang, Chen, Zhou, Xiao, and Li}]{Chen_Bao_Sun_Zhang_Chen_Zhou_Xiao_Li_2022}
Jiangjie Chen, Qiaoben Bao, Changzhi Sun, Xinbo Zhang, Jiaze Chen, Hao Zhou, Yanghua Xiao, and Lei Li. 2022.
\newblock \href {https://doi.org/10.1609/aaai.v36i10.21291} {Loren: Logic-regularized reasoning for interpretable fact verification}.
\newblock \emph{Proceedings of the AAAI Conference on Artificial Intelligence}, 36(10):10482--10491.

\bibitem[{Chen et~al.(2024{\natexlab{a}})Chen, Kim, Sriram, Durrett, and Choi}]{chen-etal-2024-complex}
Jifan Chen, Grace Kim, Aniruddh Sriram, Greg Durrett, and Eunsol Choi. 2024{\natexlab{a}}.
\newblock \href {https://doi.org/10.18653/v1/2024.naacl-long.196} {Complex claim verification with evidence retrieved in the wild}.
\newblock In \emph{Proceedings of the 2024 Conference of the North American Chapter of the Association for Computational Linguistics: Human Language Technologies (Volume 1: Long Papers)}, pages 3569--3587, Mexico City, Mexico. Association for Computational Linguistics.

\bibitem[{Chen et~al.(2023)Chen, Zhao, Zhang, Chern, Gao, Liu, and He}]{chen2023felmbenchmarkingfactualityevaluation}
Shiqi Chen, Yiran Zhao, Jinghan Zhang, I-Chun Chern, Siyang Gao, Pengfei Liu, and Junxian He. 2023.
\newblock \href {https://arxiv.org/abs/2310.00741} {Felm: Benchmarking factuality evaluation of large language models}.
\newblock \emph{Preprint}, arXiv:2310.00741.

\bibitem[{Chen et~al.(2020)Chen, Jiang, Poliak, Sakaguchi, and Van~Durme}]{chen-etal-2020-uncertain}
Tongfei Chen, Zhengping Jiang, Adam Poliak, Keisuke Sakaguchi, and Benjamin Van~Durme. 2020.
\newblock \href {https://doi.org/10.18653/v1/2020.acl-main.774} {Uncertain natural language inference}.
\newblock In \emph{Proceedings of the 58th Annual Meeting of the Association for Computational Linguistics}, pages 8772--8779, Online. Association for Computational Linguistics.

\bibitem[{Chen et~al.(2024{\natexlab{b}})Chen, Chen, Jhamtani, Xia, Shin, Eisner, and Van~Durme}]{chen-etal-2024-learning-retrieve}
Yunmo Chen, Tongfei Chen, Harsh Jhamtani, Patrick Xia, Richard Shin, Jason Eisner, and Benjamin Van~Durme. 2024{\natexlab{b}}.
\newblock \href {https://doi.org/10.18653/v1/2024.emnlp-main.406} {Learning to retrieve iteratively for in-context learning}.
\newblock In \emph{Proceedings of the 2024 Conference on Empirical Methods in Natural Language Processing}, pages 7156--7168, Miami, Florida, USA. Association for Computational Linguistics.

\bibitem[{Chern et~al.(2023)Chern, Chern, Chen, Yuan, Feng, Zhou, He, Neubig, and Liu}]{chern2023factoolfactualitydetectiongenerative}
I-Chun Chern, Steffi Chern, Shiqi Chen, Weizhe Yuan, Kehua Feng, Chunting Zhou, Junxian He, Graham Neubig, and Pengfei Liu. 2023.
\newblock \href {https://arxiv.org/abs/2307.13528} {Factool: Factuality detection in generative ai -- a tool augmented framework for multi-task and multi-domain scenarios}.
\newblock \emph{Preprint}, arXiv:2307.13528.

\bibitem[{Cho et~al.(2014)Cho, van Merri{\"e}nboer, Gulcehre, Bahdanau, Bougares, Schwenk, and Bengio}]{cho-etal-2014-learning}
Kyunghyun Cho, Bart van Merri{\"e}nboer, Caglar Gulcehre, Dzmitry Bahdanau, Fethi Bougares, Holger Schwenk, and Yoshua Bengio. 2014.
\newblock \href {https://doi.org/10.3115/v1/D14-1179} {Learning phrase representations using {RNN} encoder{--}decoder for statistical machine translation}.
\newblock In \emph{Proceedings of the 2014 Conference on Empirical Methods in Natural Language Processing ({EMNLP})}, pages 1724--1734, Doha, Qatar. Association for Computational Linguistics.

\bibitem[{Cui et~al.(2025)Cui, Yuan, Wang, Wang, Li, He, Fan, Yu, Xu, Chen, Yuan, Chen, Zhang, Lv, Wang, Yao, Han, Peng, Cheng, Liu, Sun, Zhou, and Ding}]{cui2025processreinforcementimplicitrewards}
Ganqu Cui, Lifan Yuan, Zefan Wang, Hanbin Wang, Wendi Li, Bingxiang He, Yuchen Fan, Tianyu Yu, Qixin Xu, Weize Chen, Jiarui Yuan, Huayu Chen, Kaiyan Zhang, Xingtai Lv, Shuo Wang, Yuan Yao, Xu~Han, Hao Peng, Yu~Cheng, Zhiyuan Liu, Maosong Sun, Bowen Zhou, and Ning Ding. 2025.
\newblock \href {https://arxiv.org/abs/2502.01456} {Process reinforcement through implicit rewards}.
\newblock \emph{Preprint}, arXiv:2502.01456.

\bibitem[{DeepSeek-AI(2024)}]{deepseekai2024deepseekv3technicalreport}
DeepSeek-AI. 2024.
\newblock \href {https://arxiv.org/abs/2412.19437} {Deepseek-v3 technical report}.
\newblock \emph{Preprint}, arXiv:2412.19437.

\bibitem[{Devlin et~al.(2019)Devlin, Chang, Lee, and Toutanova}]{devlin-etal-2019-bert}
Jacob Devlin, Ming-Wei Chang, Kenton Lee, and Kristina Toutanova. 2019.
\newblock \href {https://doi.org/10.18653/v1/N19-1423} {{BERT}: Pre-training of deep bidirectional transformers for language understanding}.
\newblock In \emph{Proceedings of the 2019 Conference of the North {A}merican Chapter of the Association for Computational Linguistics: Human Language Technologies, Volume 1 (Long and Short Papers)}, pages 4171--4186, Minneapolis, Minnesota. Association for Computational Linguistics.

\bibitem[{Engstrom et~al.(2019)Engstrom, Ilyas, Santurkar, Tsipras, Janoos, Rudolph, and Madry}]{engstrom2020implementationmattersdeeppolicy}
Logan Engstrom, Andrew Ilyas, Shibani Santurkar, Dimitris Tsipras, Firdaus Janoos, Larry Rudolph, and Aleksander Madry. 2019.
\newblock \href {https://arxiv.org/abs/2005.12729} {Implementation matters in deep policy gradients: A case study on ppo and trpo}.
\newblock In \emph{iclr}.

\bibitem[{Guo et~al.(2022)Guo, Schlichtkrull, and Vlachos}]{guo-etal-2022-survey}
Zhijiang Guo, Michael Schlichtkrull, and Andreas Vlachos. 2022.
\newblock \href {https://doi.org/10.1162/tacl_a_00454} {A survey on automated fact-checking}.
\newblock \emph{Transactions of the Association for Computational Linguistics}, 10:178--206.

\bibitem[{Hansen et~al.(1992)Hansen, Jaumard, and Savard}]{linearbilevelprogramming}
Pierre Hansen, Brigitte Jaumard, and Gilles Savard. 1992.
\newblock \href {https://doi.org/10.1137/0913069} {New branch-and-bound rules for linear bilevel programming}.
\newblock \emph{SIAM Journal on Scientific and Statistical Computing}, 13(5):1194--1217.

\bibitem[{Hu et~al.(2024)Hu, Long, and Wang}]{hu2024decompositiondilemmasdoesclaim}
Qisheng Hu, Quanyu Long, and Wenya Wang. 2024.
\newblock \href {https://arxiv.org/abs/2411.02400} {Decomposition dilemmas: Does claim decomposition boost or burden fact-checking performance?}
\newblock \emph{Preprint}, arXiv:2411.02400.

\bibitem[{Iqbal et~al.(2024)Iqbal, Wang, Wang, Georgiev, Geng, Gurevych, and Nakov}]{iqbal-etal-2024-openfactcheck}
Hasan Iqbal, Yuxia Wang, Minghan Wang, Georgi~Nenkov Georgiev, Jiahui Geng, Iryna Gurevych, and Preslav Nakov. 2024.
\newblock \href {https://doi.org/10.18653/v1/2024.emnlp-demo.23} {{O}pen{F}act{C}heck: A unified framework for factuality evaluation of {LLM}s}.
\newblock In \emph{Proceedings of the 2024 Conference on Empirical Methods in Natural Language Processing: System Demonstrations}, pages 219--229, Miami, Florida, USA. Association for Computational Linguistics.

\bibitem[{Jiang et~al.(2024)Jiang, Zhang, Weir, Ebner, Wanner, Sanders, Khashabi, Liu, and Durme}]{jiang2024corerobustfactualprecision}
Zhengping Jiang, Jingyu Zhang, Nathaniel Weir, Seth Ebner, Miriam Wanner, Kate Sanders, Daniel Khashabi, Anqi Liu, and Benjamin~Van Durme. 2024.
\newblock \href {https://arxiv.org/abs/2407.03572} {Core: Robust factual precision with informative sub-claim identification}.
\newblock \emph{Preprint}, arXiv:2407.03572.

\bibitem[{Kadavath et~al.(2022)Kadavath, Conerly, Askell, Henighan, Drain, Perez, Schiefer, Hatfield-Dodds, DasSarma, Tran-Johnson, Johnston, El-Showk, Jones, Elhage, Hume, Chen, Bai, Bowman, Fort, Ganguli, Hernandez, Jacobson, Kernion, Kravec, Lovitt, Ndousse, Olsson, Ringer, Amodei, Brown, Clark, Joseph, Mann, McCandlish, Olah, and Kaplan}]{kadavath2022languagemodelsmostlyknow}
Saurav Kadavath, Tom Conerly, Amanda Askell, Tom Henighan, Dawn Drain, Ethan Perez, Nicholas Schiefer, Zac Hatfield-Dodds, Nova DasSarma, Eli Tran-Johnson, Scott Johnston, Sheer El-Showk, Andy Jones, Nelson Elhage, Tristan Hume, Anna Chen, Yuntao Bai, Sam Bowman, Stanislav Fort, Deep Ganguli, Danny Hernandez, Josh Jacobson, Jackson Kernion, Shauna Kravec, Liane Lovitt, Kamal Ndousse, Catherine Olsson, Sam Ringer, Dario Amodei, Tom Brown, Jack Clark, Nicholas Joseph, Ben Mann, Sam McCandlish, Chris Olah, and Jared Kaplan. 2022.
\newblock \href {https://arxiv.org/abs/2207.05221} {Language models (mostly) know what they know}.
\newblock \emph{Preprint}, arXiv:2207.05221.

\bibitem[{Kamoi et~al.(2023)Kamoi, Goyal, Rodriguez, and Durrett}]{kamoi2023wicerealworldentailmentclaims}
Ryo Kamoi, Tanya Goyal, Juan~Diego Rodriguez, and Greg Durrett. 2023.
\newblock \href {https://arxiv.org/abs/2303.01432} {Wice: Real-world entailment for claims in wikipedia}.
\newblock \emph{Preprint}, arXiv:2303.01432.

\bibitem[{Levy et~al.(2023)Levy, Bogin, and Berant}]{levy-etal-2023-diverse}
Itay Levy, Ben Bogin, and Jonathan Berant. 2023.
\newblock \href {https://doi.org/10.18653/v1/2023.acl-long.78} {Diverse demonstrations improve in-context compositional generalization}.
\newblock In \emph{Proceedings of the 61st Annual Meeting of the Association for Computational Linguistics (Volume 1: Long Papers)}, pages 1401--1422, Toronto, Canada. Association for Computational Linguistics.

\bibitem[{Li et~al.(2024)Li, Peng, Galley, Gao, and Zhang}]{li-etal-2024-self}
Miaoran Li, Baolin Peng, Michel Galley, Jianfeng Gao, and Zhu Zhang. 2024.
\newblock \href {https://doi.org/10.18653/v1/2024.findings-naacl.12} {Self-checker: Plug-and-play modules for fact-checking with large language models}.
\newblock In \emph{Findings of the Association for Computational Linguistics: NAACL 2024}, pages 163--181, Mexico City, Mexico. Association for Computational Linguistics.

\bibitem[{Lu et~al.(2023)Lu, Qiu, Chang, Wu, Zhu, Rajpurohit, Clark, and Kalyan}]{lu2023dynamic}
Pan Lu, Liang Qiu, Kai-Wei Chang, Ying~Nian Wu, Song-Chun Zhu, Tanmay Rajpurohit, Peter Clark, and Ashwin Kalyan. 2023.
\newblock \href {https://arxiv.org/abs/2209.14610} {Dynamic prompt learning via policy gradient for semi-structured mathematical reasoning}.
\newblock In \emph{International Conference on Learning Representations (ICLR)}.

\bibitem[{Meta(2024)}]{grattafiori2024llama3herdmodels}
Meta. 2024.
\newblock \href {https://arxiv.org/abs/2407.21783} {The llama 3 herd of models}.
\newblock \emph{Preprint}, arXiv:2407.21783.

\bibitem[{Milbauer et~al.(2023)Milbauer, Ding, Wu, and Wu}]{milbauer-etal-2023-newssense}
Jeremiah Milbauer, Ziqi Ding, Zhijin Wu, and Tongshuang Wu. 2023.
\newblock \href {https://doi.org/10.18653/v1/2023.emnlp-demo.39} {{N}ews{S}ense: Reference-free verification via cross-document comparison}.
\newblock In \emph{Proceedings of the 2023 Conference on Empirical Methods in Natural Language Processing: System Demonstrations}, pages 422--430, Singapore. Association for Computational Linguistics.

\bibitem[{Min et~al.(2023)Min, Krishna, Lyu, Lewis, Yih, Koh, Iyyer, Zettlemoyer, and Hajishirzi}]{min-etal-2023-factscore}
Sewon Min, Kalpesh Krishna, Xinxi Lyu, Mike Lewis, Wen-tau Yih, Pang Koh, Mohit Iyyer, Luke Zettlemoyer, and Hannaneh Hajishirzi. 2023.
\newblock \href {https://doi.org/10.18653/v1/2023.emnlp-main.741} {{FA}ct{S}core: Fine-grained atomic evaluation of factual precision in long form text generation}.
\newblock In \emph{Proceedings of the 2023 Conference on Empirical Methods in Natural Language Processing}, pages 12076--12100, Singapore. Association for Computational Linguistics.

\bibitem[{Mnih et~al.(2016)Mnih, Badia, Mirza, Graves, Lillicrap, Harley, Silver, and Kavukcuoglu}]{pmlr-v48-mniha16}
Volodymyr Mnih, Adria~Puigdomenech Badia, Mehdi Mirza, Alex Graves, Timothy Lillicrap, Tim Harley, David Silver, and Koray Kavukcuoglu. 2016.
\newblock \href {https://proceedings.mlr.press/v48/mniha16.html} {Asynchronous methods for deep reinforcement learning}.
\newblock In \emph{Proceedings of The 33rd International Conference on Machine Learning}, volume~48 of \emph{Proceedings of Machine Learning Research}, pages 1928--1937, New York, New York, USA. PMLR.

\bibitem[{Mueller et~al.(2024)Mueller, Webson, Petty, and Linzen}]{mueller-etal-2024-context}
Aaron Mueller, Albert Webson, Jackson Petty, and Tal Linzen. 2024.
\newblock \href {https://doi.org/10.18653/v1/2024.naacl-long.267} {In-context learning generalizes, but not always robustly: The case of syntax}.
\newblock In \emph{Proceedings of the 2024 Conference of the North American Chapter of the Association for Computational Linguistics: Human Language Technologies (Volume 1: Long Papers)}, pages 4761--4779, Mexico City, Mexico. Association for Computational Linguistics.

\bibitem[{OpenAI(2022)}]{openai2022chatgpt}
OpenAI. 2022.
\newblock Chatgpt blog post.
\newblock \url{https://openai.com/index/chatgpt/}.

\bibitem[{Puterman(2014)}]{puterman2014markov}
M.L. Puterman. 2014.
\newblock \href {https://books.google.com/books?id=VvBjBAAAQBAJ} {\emph{Markov Decision Processes: Discrete Stochastic Dynamic Programming}}.
\newblock Wiley Series in Probability and Statistics. Wiley.

\bibitem[{Qiu et~al.(2021)Qiu, Yang, Ye, and Wang}]{9435807}
Shuang Qiu, Zhuoran Yang, Jieping Ye, and Zhaoran Wang. 2021.
\newblock \href {https://doi.org/10.1109/JSAIT.2021.3078754} {On finite-time convergence of actor-critic algorithm}.
\newblock \emph{IEEE Journal on Selected Areas in Information Theory}, 2(2):652--664.

\bibitem[{Raffel et~al.(2023)Raffel, Shazeer, Roberts, Lee, Narang, Matena, Zhou, Li, and Liu}]{raffel2023exploringlimitstransferlearning}
Colin Raffel, Noam Shazeer, Adam Roberts, Katherine Lee, Sharan Narang, Michael Matena, Yanqi Zhou, Wei Li, and Peter~J. Liu. 2023.
\newblock \href {https://arxiv.org/abs/1910.10683} {Exploring the limits of transfer learning with a unified text-to-text transformer}.
\newblock \emph{Preprint}, arXiv:1910.10683.

\bibitem[{Sathe et~al.(2020)Sathe, Ather, Le, Perry, and Park}]{sathe-etal-2020-automated}
Aalok Sathe, Salar Ather, Tuan~Manh Le, Nathan Perry, and Joonsuk Park. 2020.
\newblock \href {https://aclanthology.org/2020.lrec-1.849/} {Automated fact-checking of claims from {W}ikipedia}.
\newblock In \emph{Proceedings of the Twelfth Language Resources and Evaluation Conference}, pages 6874--6882, Marseille, France. European Language Resources Association.

\bibitem[{Schulman et~al.(2016)Schulman, Moritz, Levine, Jordan, and Abbeel}]{schulman2018highdimensionalcontinuouscontrolusing}
John Schulman, Philipp Moritz, Sergey Levine, Michael Jordan, and Pieter Abbeel. 2016.
\newblock \href {https://arxiv.org/abs/1506.02438} {High-dimensional continuous control using generalized advantage estimation}.
\newblock In \emph{iclr}.

\bibitem[{Schulman et~al.(2017)Schulman, Wolski, Dhariwal, Radford, and Klimov}]{schulman2017proximalpolicyoptimizationalgorithms}
John Schulman, Filip Wolski, Prafulla Dhariwal, Alec Radford, and Oleg Klimov. 2017.
\newblock \href {https://arxiv.org/abs/1707.06347} {Proximal policy optimization algorithms}.
\newblock \emph{Preprint}, arXiv:1707.06347.

\bibitem[{Schuster et~al.(2021)Schuster, Fisch, and Barzilay}]{schuster-etal-2021-get}
Tal Schuster, Adam Fisch, and Regina Barzilay. 2021.
\newblock \href {https://doi.org/10.18653/v1/2021.naacl-main.52} {Get your vitamin {C}! robust fact verification with contrastive evidence}.
\newblock In \emph{Proceedings of the 2021 Conference of the North American Chapter of the Association for Computational Linguistics: Human Language Technologies}, pages 624--643, Online. Association for Computational Linguistics.

\bibitem[{Sinha et~al.(2018)Sinha, Malo, and Deb}]{7942105}
Ankur Sinha, Pekka Malo, and Kalyanmoy Deb. 2018.
\newblock \href {https://doi.org/10.1109/TEVC.2017.2712906} {A review on bilevel optimization: From classical to evolutionary approaches and applications}.
\newblock \emph{IEEE Transactions on Evolutionary Computation}, 22(2):276--295.

\bibitem[{Song et~al.(2024)Song, Kim, and Iyyer}]{song-etal-2024-veriscore}
Yixiao Song, Yekyung Kim, and Mohit Iyyer. 2024.
\newblock \href {https://doi.org/10.18653/v1/2024.findings-emnlp.552} {{V}eri{S}core: Evaluating the factuality of verifiable claims in long-form text generation}.
\newblock In \emph{Findings of the Association for Computational Linguistics: EMNLP 2024}, pages 9447--9474, Miami, Florida, USA. Association for Computational Linguistics.

\bibitem[{Stacey et~al.(2024)Stacey, Minervini, Dubossarsky, Camburu, and Rei}]{stacey-etal-2024-atomic}
Joe Stacey, Pasquale Minervini, Haim Dubossarsky, Oana-Maria Camburu, and Marek Rei. 2024.
\newblock \href {https://doi.org/10.18653/v1/2024.emnlp-main.569} {Atomic inference for {NLI} with generated facts as atoms}.
\newblock In \emph{Proceedings of the 2024 Conference on Empirical Methods in Natural Language Processing}, pages 10188--10204, Miami, Florida, USA. Association for Computational Linguistics.

\bibitem[{Sutton and Barto(1998)}]{rl_intro}
R.S. Sutton and A.G. Barto. 1998.
\newblock \href {https://doi.org/10.1109/TNN.1998.712192} {Reinforcement learning: An introduction}.
\newblock \emph{IEEE Transactions on Neural Networks}, 9(5):1054--1054.

\bibitem[{Tang et~al.(2024)Tang, Laban, and Durrett}]{tang-etal-2024-minicheck}
Liyan Tang, Philippe Laban, and Greg Durrett. 2024.
\newblock \href {https://doi.org/10.18653/v1/2024.emnlp-main.499} {{M}ini{C}heck: Efficient fact-checking of {LLM}s on grounding documents}.
\newblock In \emph{Proceedings of the 2024 Conference on Empirical Methods in Natural Language Processing}, pages 8818--8847, Miami, Florida, USA. Association for Computational Linguistics.

\bibitem[{Thorne et~al.(2018)Thorne, Vlachos, Christodoulopoulos, and Mittal}]{thorne-etal-2018-fever}
James Thorne, Andreas Vlachos, Christos Christodoulopoulos, and Arpit Mittal. 2018.
\newblock \href {https://doi.org/10.18653/v1/N18-1074} {{FEVER}: a large-scale dataset for fact extraction and {VER}ification}.
\newblock In \emph{Proceedings of the 2018 Conference of the North {A}merican Chapter of the Association for Computational Linguistics: Human Language Technologies, Volume 1 (Long Papers)}, pages 809--819, New Orleans, Louisiana. Association for Computational Linguistics.

\bibitem[{Touvron et~al.(2023)Touvron, Lavril, Izacard, Martinet, Lachaux, Lacroix, Rozière, Goyal, Hambro, Azhar, Rodriguez, Joulin, Grave, and Lample}]{touvron2023llamaopenefficientfoundation}
Hugo Touvron, Thibaut Lavril, Gautier Izacard, Xavier Martinet, Marie-Anne Lachaux, Timothée Lacroix, Baptiste Rozière, Naman Goyal, Eric Hambro, Faisal Azhar, Aurelien Rodriguez, Armand Joulin, Edouard Grave, and Guillaume Lample. 2023.
\newblock \href {https://arxiv.org/abs/2302.13971} {Llama: Open and efficient foundation language models}.
\newblock \emph{Preprint}, arXiv:2302.13971.

\bibitem[{Vaswani et~al.(2017)Vaswani, Shazeer, Parmar, Uszkoreit, Jones, Gomez, Kaiser, and Polosukhin}]{NIPS2017_3f5ee243}
Ashish Vaswani, Noam Shazeer, Niki Parmar, Jakob Uszkoreit, Llion Jones, Aidan~N Gomez, \L~ukasz Kaiser, and Illia Polosukhin. 2017.
\newblock \href {https://proceedings.neurips.cc/paper_files/paper/2017/file/3f5ee243547dee91fbd053c1c4a845aa-Paper.pdf} {Attention is all you need}.
\newblock In \emph{Advances in Neural Information Processing Systems}, volume~30. Curran Associates, Inc.

\bibitem[{Wang et~al.(2022)Wang, Mishra, Alipoormolabashi, Kordi, Mirzaei, Naik, Ashok, Dhanasekaran, Arunkumar, Stap, Pathak, Karamanolakis, Lai, Purohit, Mondal, Anderson, Kuznia, Doshi, Pal, Patel, Moradshahi, Parmar, Purohit, Varshney, Kaza, Verma, Puri, Karia, Doshi, Sampat, Mishra, Reddy~A, Patro, Dixit, and Shen}]{wang-etal-2022-super}
Yizhong Wang, Swaroop Mishra, Pegah Alipoormolabashi, Yeganeh Kordi, Amirreza Mirzaei, Atharva Naik, Arjun Ashok, Arut~Selvan Dhanasekaran, Anjana Arunkumar, David Stap, Eshaan Pathak, Giannis Karamanolakis, Haizhi Lai, Ishan Purohit, Ishani Mondal, Jacob Anderson, Kirby Kuznia, Krima Doshi, Kuntal~Kumar Pal, Maitreya Patel, Mehrad Moradshahi, Mihir Parmar, Mirali Purohit, Neeraj Varshney, Phani~Rohitha Kaza, Pulkit Verma, Ravsehaj~Singh Puri, Rushang Karia, Savan Doshi, Shailaja~Keyur Sampat, Siddhartha Mishra, Sujan Reddy~A, Sumanta Patro, Tanay Dixit, and Xudong Shen. 2022.
\newblock \href {https://doi.org/10.18653/v1/2022.emnlp-main.340} {Super-{N}atural{I}nstructions: Generalization via declarative instructions on 1600+ {NLP} tasks}.
\newblock In \emph{Proceedings of the 2022 Conference on Empirical Methods in Natural Language Processing}, pages 5085--5109, Abu Dhabi, United Arab Emirates. Association for Computational Linguistics.

\bibitem[{Wang et~al.(2024)Wang, Reddy, Mujahid, Arora, Rubashevskii, Geng, Afzal, Pan, Borenstein, Pillai, Augenstein, Gurevych, and Nakov}]{wang2024factcheckbenchfinegrainedevaluationbenchmark}
Yuxia Wang, Revanth~Gangi Reddy, Zain~Muhammad Mujahid, Arnav Arora, Aleksandr Rubashevskii, Jiahui Geng, Osama~Mohammed Afzal, Liangming Pan, Nadav Borenstein, Aditya Pillai, Isabelle Augenstein, Iryna Gurevych, and Preslav Nakov. 2024.
\newblock \href {https://arxiv.org/abs/2311.09000} {Factcheck-bench: Fine-grained evaluation benchmark for automatic fact-checkers}.
\newblock \emph{Preprint}, arXiv:2311.09000.

\bibitem[{Wanner et~al.(2024)Wanner, Ebner, Jiang, Dredze, and Durme}]{wanner2024closerlookclaimdecomposition}
Miriam Wanner, Seth Ebner, Zhengping Jiang, Mark Dredze, and Benjamin~Van Durme. 2024.
\newblock \href {https://arxiv.org/abs/2403.11903} {A closer look at claim decomposition}.
\newblock \emph{Preprint}, arXiv:2403.11903.

\bibitem[{Wei et~al.(2024)Wei, Yang, Song, Lu, Hu, Huang, Tran, Peng, Liu, Huang, Du, and Le}]{wei2024longformfactualitylargelanguage}
Jerry Wei, Chengrun Yang, Xinying Song, Yifeng Lu, Nathan Hu, Jie Huang, Dustin Tran, Daiyi Peng, Ruibo Liu, Da~Huang, Cosmo Du, and Quoc~V. Le. 2024.
\newblock \href {https://arxiv.org/abs/2403.18802} {Long-form factuality in large language models}.
\newblock \emph{Preprint}, arXiv:2403.18802.

\bibitem[{Zhang et~al.(2022)Zhang, Feng, and Tan}]{zhang-etal-2022-active}
Yiming Zhang, Shi Feng, and Chenhao Tan. 2022.
\newblock \href {https://doi.org/10.18653/v1/2022.emnlp-main.622} {Active example selection for in-context learning}.
\newblock In \emph{Proceedings of the 2022 Conference on Empirical Methods in Natural Language Processing}, pages 9134--9148, Abu Dhabi, United Arab Emirates. Association for Computational Linguistics.

\end{thebibliography}

\onecolumn
\clearpage
\appendix

\begin{center}
{\LARGE \textbf{Supplemental Material}}
\end{center}

\begin{center}
\begin{tabular}{@{}ll@{}}
\toprule
\textbf{Appendix} & \textbf{Contents} \\ \midrule
\autoref{appendix: limitations of decomposition policies} & Limitations of Static Decomposition Policies \\
\autoref{appendix: experiment setup}  & Additional Details of Experimental Setup  \\
\autoref{appendix: experiment results} & Additional Details of Experimental Results \\
\autoref{appendix: prompts}       &   Used Prompts\\ 
\autoref{appendix: example} & Example \\
\bottomrule
\end{tabular}
\end{center}

\section{Limitations of Static Decomposition Policies}
\label{appendix: limitations of decomposition policies}
\paragraph{Static decomposition demonstrations hardly handle input with varying fact density.} \citet{song-etal-2024-veriscore} find that the factual score evaluated on one task (e.g., biography generation) does not necessarily correlate with the one evaluated on different tasks (e.g., long-form QA), which have different input atomicities.
Similar issues have also been revealed in other compositional tasks, such as semantic parsing, where static ICL demonstrations have unreliable performance on out-of-distribution data \citep{levy-etal-2023-diverse, mueller-etal-2024-context}. Thus, existing decomposition policies that rely on static demonstrations, including FActScore \citep{min-etal-2023-factscore}, WICE \citep{kamoi2023wicerealworldentailmentclaims}, R-ND \citep{wanner2024closerlookclaimdecomposition}, and SAFE \cite{wei2024longformfactualitylargelanguage}, may struggle to handle input claims with varying fact densities or atomicities.

\paragraph{Subclaims are atomically homogeneous and not optimized for downstream verifiers.} There is a prevailing assumption made in prior works that verification performance is expected to increase as input complexity decreases \citep{min-etal-2023-factscore, wei2024longformfactualitylargelanguage, hu2024decompositiondilemmasdoesclaim}. Therefore, they leverage in-context demonstrations to establish uniformly low atomicity among all generated subclaims, which barely accommodate downstream verifiers based on our findings.

Our \method{} is the first solution to tackle these limitations by explicitly exploring various decomposition complexities during training. As a result, it achieves a more generalizable performance and consistently outperforms static methods across varying input atomicities.

\section{Experiment Setup}
\label{appendix: experiment setup}

\subsection{Data Preparation}
\label{appendix: dataset statistics}
We first split the given FActScore dataset into train (60\%), validation (20\%), and test (20\%) sets. Subclaims are then recursively constructed,\footnote{We use the same protocol as in Eq.\ref{eq: objective} to label new subclaims: a claim is true only if all its subclaims are true.} with each atomicity level having its own train, validation, and test sets. We present the number of claims per atomicity in \autoref{table: data statistics}.
\begin{table}[ht]
\centering
\begin{tabular}{@{}cccccccc@{}}
\toprule
\multirow{2}{*}{\begin{tabular}[c]{@{}l@{}}\textbf{Atomicity}$\rightarrow$\\ \textbf{Data Source}$\downarrow$\end{tabular}} & \multicolumn{1}{c}{\multirow{2}{*}{4}} & \multicolumn{1}{c}{\multirow{2}{*}{3}} & \multicolumn{1}{c}{\multirow{2}{*}{2}} & \multicolumn{1}{c}{\multirow{2}{*}{1}} & \multicolumn{1}{c}{\multirow{2}{*}{0}} & \multicolumn{1}{c}{\multirow{2}{*}{-1}} \\
                                                                                 & \multicolumn{1}{c}{}                   & \multicolumn{1}{c}{}                   & \multicolumn{1}{c}{}                   & \multicolumn{1}{c}{}                   & \multicolumn{1}{c}{}                   & \multicolumn{1}{c}{}                    \\ \midrule
ChatGPT                                                                          & 89                                     & 251                                    & 578                                    & 1239                                   & 4886                                   & 2482                                    \\
PerplexityAI                                                                     & 116                                    & 289                                    & 661                                    & 1405                                   & 5568                                   & 2842                                    \\ \bottomrule
\end{tabular}
\caption{Statistics of constructed subclaim datasets.}
\label{table: data statistics}
\end{table}

Note that we do not use subclaims with atomicity -1 during training because they are already over-decomposed. Thus only the test set at atomicity 0 is decomposed to get subclaims at $-1$, resulting in fewer instances at atomicity -1 compared to at 0.

\subsection{Data Example}
\label{appendix: data example}
\begin{tcolorbox}[colback=gray!5!white,colframe=gray!75!black,title=\textbf{Data from ChatGPT}]
\tt \small
Quintus Sosius Senecio was a Roman senator and military commander who lived during the 2nd century AD. He was a member of the senatorial class and served in various high-ranking positions in the Roman government, including as consul in 128 AD.
\\ \\
In addition to his political career, Quintus Sosius Senecio was also a distinguished military leader. He served as a legate (commander) in several provinces, including Africa and Pannonia, and was awarded triumphal honors for his victories over the enemies of Rome.
\\ \\
Quintus Sosius Senecio was also known for his literary works, including a history of the Roman Empire, which unfortunately has not survived to the present day.
\\ \\
Overall, Quintus Sosius Senecio was a prominent figure in Roman society, known for his military prowess, political acumen, and intellectual pursuits.
\end{tcolorbox}

\begin{tcolorbox}[colback=gray!5!white,colframe=gray!75!black,title=\textbf{Data from PerplexityAI}]
\tt \small
Quintus Sosius Senecio was a Roman consul who lived during the 1st and 2nd centuries AD[1]. He was married to Julia, the daughter of Frontinus, a prominent Roman civil engineer, author, soldier, and senator[3]. Quintus Sosius Senecio was the father of Sosia Polla[1], who married Quintus Pompeius Falco, a consul in AD 109[4].
\\ \\
Quintus Sosius Senecio was a friend of Plutarch, a Greek Middle Platonist philosopher, historian, biographer, essayist[2]. Plutarch wrote about several Roman nobles in his works including Quintus Sosius Senecio and Titus Avidius Quietus[2].
\end{tcolorbox}

\subsection{Training Initialization}
\label{appendix: training initialization}
We initialize our atomicity state as a zero vector with dimension size 768 and set the bias of the update gate in GRU to $-\infty$. This is because \citet{chen-etal-2024-learning-retrieve} found that using an identity function in state transition helps stabilize RL training.
We use BERT \citep{devlin-etal-2019-bert} to obtain embeddings for Eq.\ref{eq: state transition}, where subclaims are concatenated using the \texttt{[SEP]} token.\footnote{\href{https://huggingface.co/google-bert/bert-base-uncased}{https://huggingface.co/google-bert/bert-base-uncased}} We employ an UNLI model \citep{chen-etal-2020-uncertain} to estimate the conditional probability $P(\cdot \mid \mathcal{H})$ in Eq.\ref{eq: information loss}.\footnote{\href{https://huggingface.co/Zhengping/roberta-large-unli}{https://huggingface.co/Zhengping/roberta-large-unli}} Following \citet{jiang2024corerobustfactualprecision}, we use these bleached contextual claims showed in \autoref{table: bleached claim set} as $\mathcal{H}$.

\begin{table}[ht]
    \centering
    \begin{tabular}{c}
        \toprule
        \textbf{Claim Template}\\
        \midrule
          {\fontfamily{qcr}\selectfont \$\{TOPIC\}} is a person. \\
          {\fontfamily{qcr}\selectfont \$\{TOPIC\}} breathes.\\
          {\fontfamily{qcr}\selectfont \$\{TOPIC\}} exists.\\
          {\fontfamily{qcr}\selectfont \$\{TOPIC\}} is a name.\\
          {\fontfamily{qcr}\selectfont \$\{TOPIC\}} is unique.\\
          {\fontfamily{qcr}\selectfont \$\{TOPIC\}} is famous.\\
          {\fontfamily{qcr}\selectfont \$\{TOPIC\}} has some abilities.\\
          somebody knows {\fontfamily{qcr}\selectfont \$\{TOPIC\}}.\\
          {\fontfamily{qcr}\selectfont \$\{TOPIC\}} is a star.\\
        \bottomrule
    \end{tabular}
    \caption{Bleached claim set designed for FActScore-style biography evaluation.}
    \label{table: bleached claim set}
\end{table}

We provide our binary decomposition prompt in Appendix \ref{appendix: prompts}. Similar to \citep{cui2025processreinforcementimplicitrewards}, we employ online trajectory filtering, which filters out trajectories based on a predefined reward mean threshold. 

\subsection{Hyperparameters}
\label{appendix: hyperparameters}
For decomposition, we set the sampling temperature to 0.2 during training to encourage exploration and use 0 during evaluation for better reproducibility. We consistently set the temperature to 0 for verification. 

For PPO training, we use the following hyperparameters: $\epsilon=0.2$, $\gamma=0.99$, $\lambda=0.95$, $c_1=0.02$, $c_2=0.005$. We configure the replay buffer size to 512 steps, the rollout batch size to 32 samples, the maximum decomposition trajectory length to 20 steps, the mini-batch size to 32, and the trajectory filter threshold to -0.02. We use the learning rate $3e^{-5}$ together with a cosine-annealing learning rate scheduler. The model is trained for 100 steps, with validation performed every 10 steps, and the best-performing model is saved. Training is conducted on 2 NVIDIA RTX A6000 GPUs, requiring approximately 80 GPU hours.

\newpage
\section{Experiment Results}
\label{appendix: experiment results}

\begin{figure}[ht]
\centering
\begin{subfigure}[b]{1\textwidth}
   \includegraphics[width=1\linewidth]{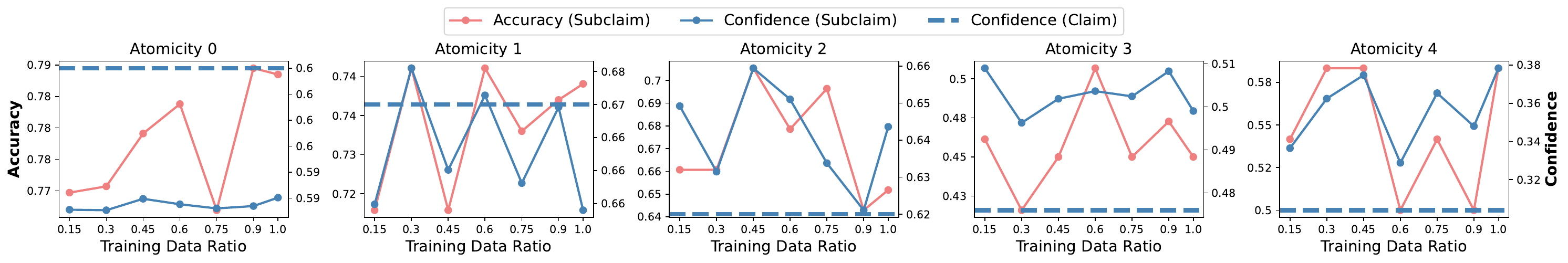}
   \caption{Verification results on ChatGPT dataset}
\end{subfigure}

\begin{subfigure}[b]{1\textwidth}
   \includegraphics[width=1\linewidth]{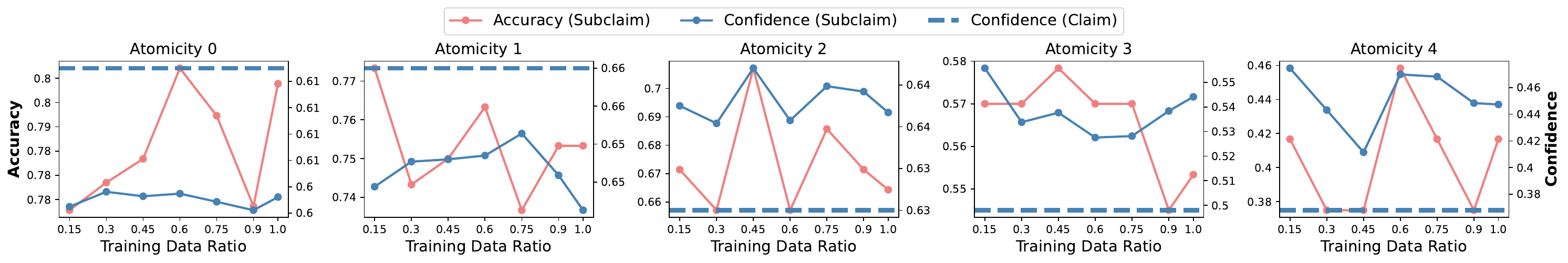}
   \caption{Verification results on PerplexityAI dataset}
\end{subfigure}
\caption{The verification sensitivity of \method{} as the training data size changes. The five figures (from left to right) represent claims with atomicity in the range $[0,4]$, evaluated under \short{} policy trained on different dataset sizes. The horizontal dashed line denotes verification confidence for original claims without decomposition. We use decomposition LLM \texttt{Llama3-Inst-70B} and verification LLM \texttt{Llama3-Inst-8B} with retrieval verification policy.}
\label{fig: ablation study}
\end{figure}

\paragraph{How much data is needed for training a \method{} policy?}
Given \short{} requires iterative online practice to determine optimal atomicity, it is natural to ask how much data is required to train an effective \short{} policy. The results are presented in \autoref{fig: ablation study}, where we trained six \short{} policies using varying amounts of training data. A training data ratio of $1$ represents the dataset size used in our main experiment. 
We observe that \textbf{\short{} with on-policy learning can achieve promising and comparable performance even with a limited amount of data}. Generally, \short{} performance improves with increased training data, particularly on claims with lower atomicity, but begins to oscillate after reaching a certain data threshold (e.g. 0.75). 

\begin{figure}[ht]
    \centering
    \includegraphics[width=0.7\linewidth]{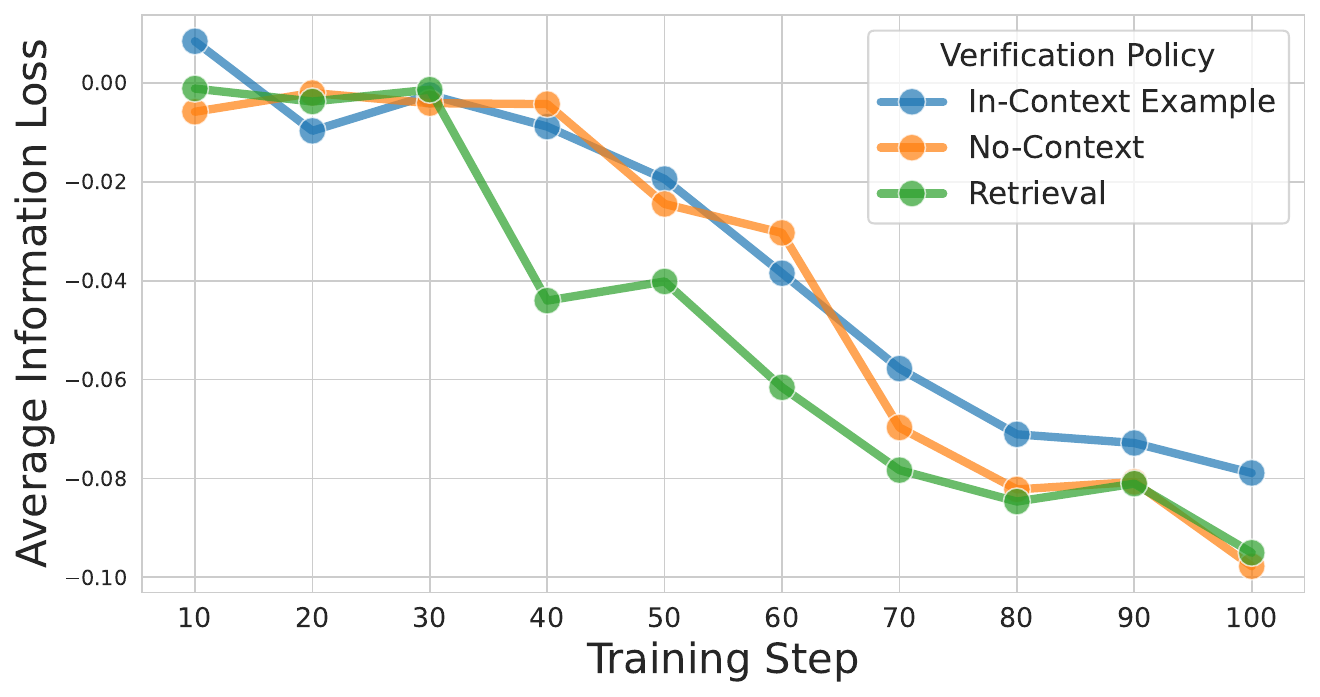}
    \caption{Average information loss measures across training steps on the validation set. Clearly, it decreases as the model continues training and learning to decompose claims into more atomic levels. This trend aligns with our design motivation outlined in Eq.\ref{eq: information loss}: as claims become syntactically closer to bleached claims through decomposition, the resulting information loss diminishes accordingly.}
    \label{fig: avg info loss}
\end{figure}

\setlength{\aboverulesep}{2.9pt}
\setlength{\belowrulesep}{2.9pt}
\begin{table*}[ht]
\centering
\small
\begin{tabular}{@{}>{\centering\arraybackslash}p{1.2cm}|p{2.7cm}ccccc@{}}
\toprule
\multirow{2}{*}{\textbf{Atomicity}} &
\multirow{2}{*}{\begin{tabular}[c]{@{}l@{}}
  \textbf{Decompose Policy} $\rightarrow$\\
  \textbf{Verify Policy}$\downarrow$
\end{tabular}} & FActScore & FActScore-Atom & WICE & R-ND & \textbf{\short} \\
& & \multicolumn{5}{c}{\textbf{\textit{Verification Confidence [0-1] $\uparrow$ / Verification Accuracy [0-1] $\uparrow$}}} \\
\midrule
\multirow{4}{*}{0}
  & \multicolumn{6}{l}{\textit{Data Source: PerplexityAI}} \\
  & Retrieval  & 0.702 / 0.844 & 0.655 / 0.770 & 0.705 / 0.856 & 0.700 / 0.802 & 0.685 / 0.820 \\
  & In-Context Example & 0.741 / 0.309 & 0.702 / 0.197 & 0.757 / 0.334 & 0.730 / 0.283 & \textbf{0.757} / 0.323 \\
  & No-Context &0.593 / 0.522 & 0.557 / 0.454 & 0.600 / 0.553 & 0.600 / 0.463 & 0.580 / 0.507 \\
\midrule
\multirow{4}{*}{1}
  & \multicolumn{6}{l}{\textit{Data Source: PerplexityAI}} \\
  & Retrieval  & 0.694 / 0.740 & 0.685 / 0.717  & 0.690 / 0.760 & 0.681 / 0.673 & 0.680 / \textbf{0.780} \\
  & In-Context Example & 0.754 / 0.260 & 0.743 / 0.247 & 0.763 / 0.273 & 0.749 / 0.230 & \textbf{0.768} / \textbf{0.333} \\
  & No-Context  & 0.583 / 0.310 & 0.561 / 0.337 & 0.563 / 0.347 & 0.591 / 0.290 & 0.569 / \textbf{0.457}\\
\midrule
\multirow{4}{*}{2} 
  & \multicolumn{6}{l}{\textit{Data Source: PerplexityAI}} \\
  & Retrieval          & 0.706 / 0.692 & 0.695 / 0.650 & 0.700 / 0.671 & 0.675 / 0.622 & 0.656 / 0.679\\
  & In-Context Example & 0.749 / 0.385 & 0.748 / 0.385 & 0.763 / 0.385  & 0.747 / 0.385 & \textbf{0.802} / \textbf{0.400} \\
  & No-Context         & 0.594 / 0.364  & 0.586 / 0.385 & 0.586 / 0.371  & 0.587 / 0.385 & 0.564 / \textbf{0.386} \\
\bottomrule
\end{tabular}
\caption{Comparison of our \short{} over baselines on the testing set. Each metric is scaled from 0 to 1. $\uparrow$ indicates higher values are preferred. We employ decomposition LLM \texttt{DeepSeek-V3} and verification LLM \texttt{Llama3-Inst-8B}.
\textbf{\short{} consistently outperforms baselines on verification accuracy across various verifiers and input claim atomicities.}}
\label{table: comparison results ds-v3}
\end{table*}

\newpage
\section{Prompts}
\label{appendix: prompts}

\begin{tcolorbox}[colback=gray!5!white,colframe=gray!75!black,title=\textbf{Binary Decomposition}]
\tt \small
[system]
You are a decomposer. Your task is to decompose the given claim into two sub-claims. There are two principles you have to follow: 1) making sure there is no information loss or gain after decomposition and 2) making sure each generated subclaim is self-contained and approximately equal in length and information. Seperate the two subclaims with a hyphen. \\

[user]
Following the given two principles, please decompose the following claim into two sub-claims: In 1963, Collins became one of the third group of astronauts selected by NASA and he served as the back-up Command Module Pilot for the Gemini 7 mission.\\
- Collins became one of the third group of astronauts selected by NASA in 1963.\\
- Collins served as the back-up Command Module Pilot for the Gemini 7 mission.\\

Following the given two principles, please decompose the following claim into two sub-claims: In addition to his acting roles, Bateman has written and directed two short films and is currently in development on his feature debut.\\
- In addition to his acting roles, Bateman has written and directed two short films.\\
- Bateman is currently in development on his feature debut.\\

Following the given two principles, please decompose the following claim into two sub-claims: "Parasite" received widespread critical acclaim for its screenplay, direction, acting, and its social commentary.\\
- "Parasite" received widespread critical acclaim for its screenplay and direction.\\
- "Parasite" received widespread critical acclaim for its acting and social commentary.\\

Following the given two principles, please decompose the following claim into two sub-claims: \{claim\}

\end{tcolorbox}

\begin{tcolorbox}[colback=gray!5!white,colframe=gray!75!black,title=\textbf{FActScore-Atom Decomposition Policy}]
\tt \small
[system]
You are a decomposer. Your task is to decompose the given claim into more granular subclaims. There are two principles you have to follow: 1) making sure there is no information loss or gain after decomposition and 2) making sure each generated subclaim is self-contained. Seperate the decomposed subclaims with a hyphen \\

[user]
Following the given two principles, please decompose the following claim into more granular subclaims: He made his acting debut in the film The Moon is the Sun's Dream.\\
- He made his acting debut.\\
- Debut happened in the film.\\
- The Moon is the Sun's Dream is a film.\\

Following the given two principles, please decompose the following claim into more granular subclaims: He has worked with a wide variety of artists.\\
- He worked.\\
- It happened with a wide variety of artists.\\

Following the given two principles, please decompose the following claim into more granular subclaims: Bateman has directed two short films.\\
- Bateman had directed films.\\
- There are two films.\\
- Films are short.\\

Following the given two principles, please decompose the following claim into more granular subclaims: \{claim\}
\end{tcolorbox}

\begin{tcolorbox}[colback=gray!5!white,colframe=gray!75!black,title=\textbf{In-Context Examples Verification Policy}]
\tt\small
[system]
Your task is to verify the correctness of the given claim. Only answer with 'True' or 'False'.\\

[user]
Input: In 1996, he was awarded the Ig Nobel Prize for Art, an award given to honor achievements that make people laugh, and then think. True or False?\\
Output: True\\
Input: Featherstone's pink flamingo design was displayed at the Smithsonian National Museum of American History in 1996, and he was inducted into the Plastics Hall of Fame in 1998. True or False?\\
Output: False\\
Input: Featherstone continued to work on his designs until his death in 2015, and his creations remain popular among collectors and enthusiasts of Americana. True or False?\\
Output: False\\
Input: Travis Oliphant is a data scientist and entrepreneur who is best known for creating the NumPy and SciPy libraries for Python programming language. True or False?\\
Output: True\\
Input: He was born on August 22, 1972, in the United States. True or False?\\
Output: False
\end{tcolorbox}

\newpage
\section{Example}
\label{appendix: example}

\begin{table}[H]
\footnotesize
\centering
\resizebox{.99\textwidth}{!}{
\begin{tabular}{p{0.1\textwidth}p{0.85\textwidth}}
\toprule
\begin{tabular}{p{0.1\textwidth}}\textbf{Method}\end{tabular}& \begin{tabular}{p{0.85\textwidth}}\textbf{ Results} \end{tabular}\\\midrule
\\
\begin{tabular}{p{0.1\textwidth}}
Original \end{tabular} & 
\tt
\begin{tabular}{p{0.85\textwidth}} 
After retiring from tennis in 1996, Sabatini pursued other interests, including fashion and fragrance design. She also established the Gabriela Sabatini Foundation, which supports children's education and sports programs in Argentina.
\end{tabular} \\&\\
\cdashline{1-2}
\begin{tabular}{p{0.1\textwidth}}
FactScore \citep{min-etal-2023-factscore}\end{tabular} & 
\tt
\begin{tabular}{p{0.85\textwidth}} 
\begin{enumerate}
\itemsep0em
\item Sabatini retired from tennis in 1996.
\item After retiring from tennis, Sabatini pursued other interests.
\item Sabatini pursued fashion design.
\item Sabatini pursued fragrance design.
\item Sabatini established the Gabriela Sabatini Foundation.
\item The Gabriela Sabatini Foundation supports children's education.
\item The Gabriela Sabatini Foundation supports sports programs.
\item The Gabriela Sabatini Foundation supports children's education and sports programs in Argentina.
\end{enumerate}
\end{tabular} \\&\\
\cdashline{1-2}
\begin{tabular}{p{0.1\textwidth}}
R-ND \citep{wanner2024closerlookclaimdecomposition} \end{tabular} & 
\tt
\begin{tabular}{p{0.85\textwidth}} 
\begin{enumerate}
\itemsep0em
\item Sabatini retired from tennis. 
\item Sabatini's retirement from tennis occurred in 1996.
\item Sabatini pursued other interests after retiring.
\item Sabatini's other interests include fashion. 
\item Sabatini's other interests include fragrance design.
\item Sabatini established the Gabriela Sabatini Foundation.
\item The Gabriela Sabatini Foundation supports children's education.
\item The Gabriela Sabatini Foundation supports sports programs.
\item The Gabriela Sabatini Foundation operates in Argentina.
\item Sabatini's establishment of the foundation occurred after her retirement.
\item Sabatini's pursuit of other interests occurred after her retirement.
\end{enumerate}
\end{tabular} \\&\\
\cdashline{1-2}
\begin{tabular}{p{0.1\textwidth}}
WICE \citep{kamoi2023wicerealworldentailmentclaims} \end{tabular} & 
\tt
\begin{tabular}{p{0.85\textwidth}} 
\begin{enumerate}
\itemsep0em
\item Sabatini retired from tennis in 1996.
\item Sabatini pursued other interests after retiring.
\item Sabatini's interests included fashion design.
\item Sabatini's interests included fragrance design.
\item Sabatini established the Gabriela Sabatini Foundation.
\item The Gabriela Sabatini Foundation supports children's education in Argentina.
\item The Gabriela Sabatini Foundation supports sports programs in Argentina.
\end{enumerate}
\end{tabular}
\\&\\
\cdashline{1-2}
\begin{tabular}{p{0.1\textwidth}}
\textbf{\short{} (Ours)} \end{tabular} & 
\tt
\begin{tabular}{p{0.85\textwidth}} 
\begin{enumerate}
\itemsep0em
\item After retiring from tennis in 1996, Sabatini pursued other interests, including fashion and fragrance design.
\item Sabatini established the Gabriela Sabatini Foundation, which supports children's education and sports programs in Argentina.
\end{enumerate}
\end{tabular}
\\ \bottomrule
\end{tabular}
}
\caption{Example of decomposition results from different methods, where we use \texttt{DeepSeek-V3} as the decomposer. Our \short{} policy is trained for the verifier \texttt{Llama3-Inst-8B} with No-Context policy.}
\label{table: decomposition example}
\end{table}

We provide an example in \autoref{table: decomposition example} to show the differences between subclaims decomposed using our \short{} policy and those derived from popular decomposition methods. For the verifier \texttt{Llama3-Inst-8B} using No-Context verification policy, which according to \autoref{fig: granularity preference} has the best verification performance at atomicity level 2 (i.e., input contains 4 pieces of atomic information). Evidently, only our trained \short{} policy successfully generates subclaims that closely match the verifier's preferred atomicity level. These examples highlight how existing decomposition methods can lead to suboptimal performance in fact-checking systems.

\end{document}